\documentclass[lettersize,journal]{IEEEtran}
\usepackage{amsmath,amsfonts}
\usepackage{algorithmic}
\usepackage{algorithm}
\usepackage{array}
\usepackage[caption=false,font=normalsize,labelfont=sf,textfont=sf]{subfig}
\usepackage{textcomp}
\usepackage{stfloats}
\usepackage{url}
\usepackage{verbatim}
\usepackage{graphicx}
\usepackage{cite}
\usepackage{multirow}
\usepackage[table,xcdraw]{xcolor}
\usepackage[normalem]{ulem}
\useunder{\uline}{\ul}{}

\begin{document}

\title{Dual Mutual Learning Network with Global–local Awareness for RGB-D Salient Object Detection}

\author{Kang Yi, Haoran Tang, Yumeng Li, Jing Xu, Jun Zhang

\thanks{This work is supported by the National Natural Science Foundation of China (Grant No. 62233011), Tianjin Natural Science Foundation, China (Grant No. 21JCYBJC00110). (Kang Yi and Haoran Tang contributed equally to this work.) (Corresponding author: Jing Xu.) 

Yi Kang, Yumeng Li, Jing Xu and Jun Zhang are with the College of Artificial Intelligence, Nankai University, Tianjin 300350, China. Haoran Tang is with the Department of Computing, The Hong Kong Polytechnic University, Kowloon, Hong Kong.
}
}

\markboth{Submitted to IEEE Transcation on Multimedia}%
{Shell \MakeLowercase{\textit{et al.}}: A Sample Article Using IEEEtran.cls for IEEE Journals}

\maketitle

\begin{abstract}
RGB-D salient object detection (SOD), aiming to highlight prominent regions of a given scene by jointly modeling RGB and depth information, is one of the challenging pixel-level prediction tasks. Recently, the dual-attention mechanism has been devoted to this area due to its ability to strengthen the detection process. However, most existing methods directly fuse attentional cross-modality features under a manual-mandatory fusion paradigm without considering the inherent discrepancy between the RGB and depth, which may lead to a reduction in performance. Moreover, the long-range dependencies derived from global and local information make it difficult to leverage a unified efficient fusion strategy. Hence, in this paper, we propose the GL-DMNet, a novel dual mutual learning network with global-local awareness. Specifically, we present a position mutual fusion module and a channel mutual fusion module to exploit the interdependencies among different modalities in spatial and channel dimensions. Besides, we adopt an efficient decoder based on cascade transformer-infused reconstruction to integrate multi-level fusion features jointly. Extensive experiments on six benchmark datasets demonstrate that our proposed GL-DMNet performs better than 24 RGB-D SOD methods, achieving an average improvement of $\sim$3\% across four evaluation metrics compared to the second-best model (S3Net). Codes and results are available at https://github.com/kingkung2016/GL-DMNet.
\end{abstract}


\section{Introduction}
\IEEEPARstart{S}{alient} object detection (SOD) is one of the most fundamental yet challenging problems in computer vision, whose goal is to locate the most visually attractive objects or regions of a given scene \cite{SIP_D3Net2021TNNLS,zhou2021rgb}. It has been successfully applied in various fields, such as autonomous driving \cite{wu2024s}, person identification \cite{dreamt}, image editing \cite{li2023lightweight}, and medical image understanding \cite{zhang2021brain}. Nevertheless, the conventional methods for SOD task face difficulties when handling complex and indistinguishable scenarios, causing suboptimal performance. Especially in the real world, there is rich RGB and depth information (RGB-D). Recently, the study on dual attention mechanism, which could effectively capture feature dependencies in the spatial and channel dimensions \cite{fu2019dual}, provides a new opportunity to this area. Thus, it is possible to improve the performance of RGB-D SOD across different scenarios without constraints.

\begin{figure}[!t]
	\centering
	\includegraphics[width=0.48\textwidth]{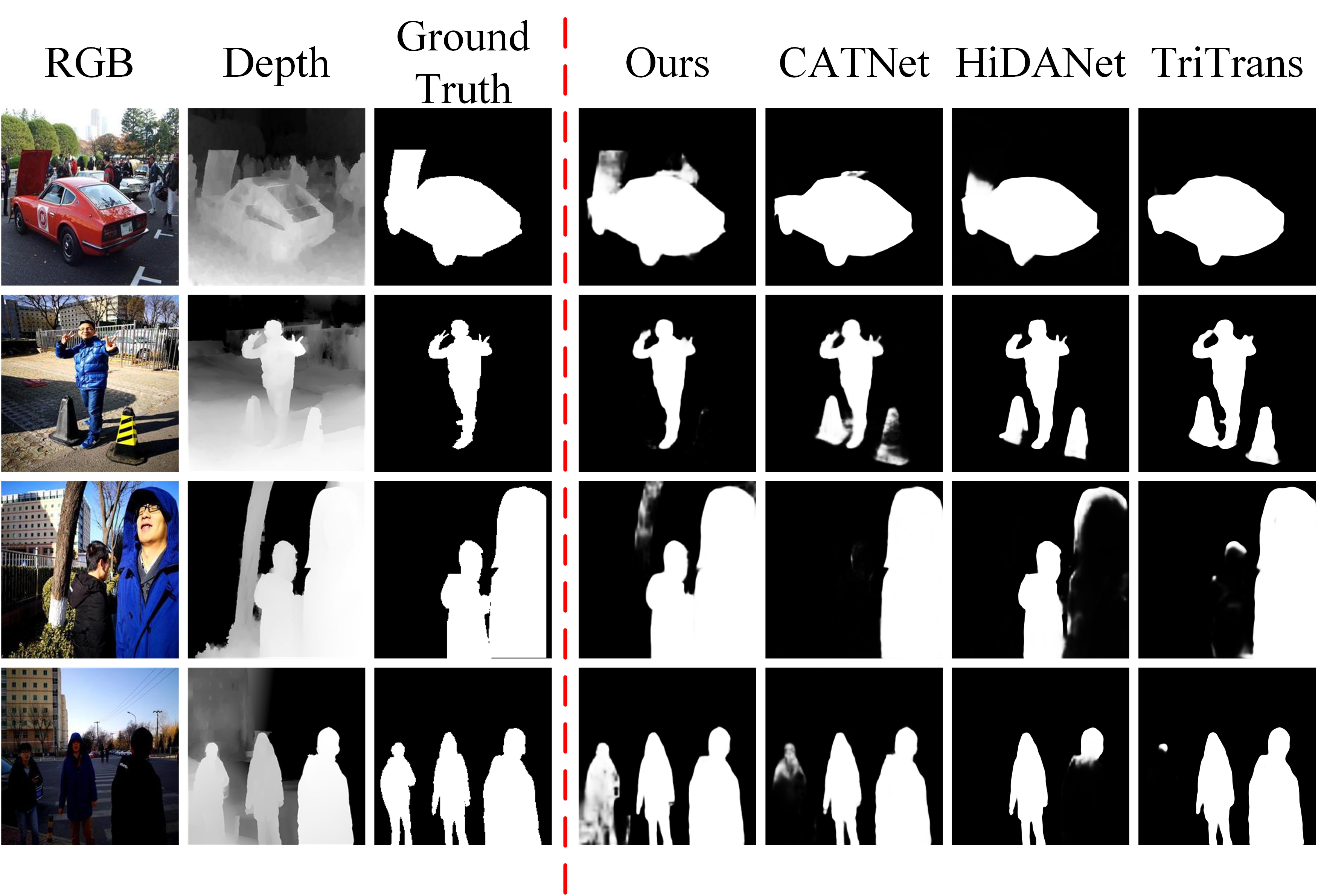}
	\centering
	\caption{The results of our GL-DMNet and other representative methods, including CATNet \cite{CATNet}, HiDANet \cite{wu2023hidanet} and TriTransNet \cite{liu2021tritransnet}.}
	\label{fig:examples}
\end{figure}

The idea of dual attention is to model the long-range contextual dependencies by dual attention modules \cite{fu2019dual}, e.g., position and channel attentions. Consequently, it exploits more detailed semantic information from the high-level features of RGB images and depth images and enhances them in turn \cite{fu2021scene, CATNet}. However, previous methods directly integrate different attention modules under the manual-mandatory fusion strategy to conduct multi-modal feature learning, neglecting the inherent discrepancy (e.g., semantic content) between the RGB and the depth \cite{fang2023m2rnet}. As a result, most of them may need more flexibility in multi-modal feature fusion, which further limits performance improvement. Hence, efficiently investigating and combining multi-modal information becomes an urgent issue for the RGB-D SOD task.

Another issue occurring in RGB-D SOD is the deficiency of collaborative effect between local correlation and global correlation for each pixel \cite{liu2021tritransnet}. Especially for multi-modal learning, the RGB and depth features would pose more long-range dependencies \cite{bi2023cross, zeng2024airsod}, making it hard for the RGB and depth features to complement each other. As a consequence, the perception of global-local awareness may be destroyed passively. Thus, sufficiently capturing cross-modality global-local context is another challenge that needs to be solved.

To tackle the above issues, we propose a novel model, the dual mutual learning network with global-local awareness (GL-DMNet) for the RGB-D SOD task. Firstly, to fully investigate the features of RGB and depth information, we incorporate the dual attention mechanism into cross-modality learning, which naturally connects different local features and adaptively integrates local features with their global dependencies. By applying the dual attention mechanism, our model could learn cross-modality features at any scale from a global view. In other words, we endow our framework with the capabilities of local and global learning. As a result, we would obtain multi-scale complementary information, and then we can devise multi-modal context enhancement to improve the quality of the SOD task. Secondly, we design an interactive manner by constructing a joint transformer-based and CNN-based encoder. This innovation allows us to globally capture the correlation between different stages, further mitigating the previously-mentioned issues. Last, we leverage an efficient decode strategy that follows the transformed-infused reconstruction to accurately conduct our final SOD predictions and also free the whole model from heavy fusion learning.

\begin{figure}[ht]
	\centering
	\includegraphics[width=0.485\textwidth]{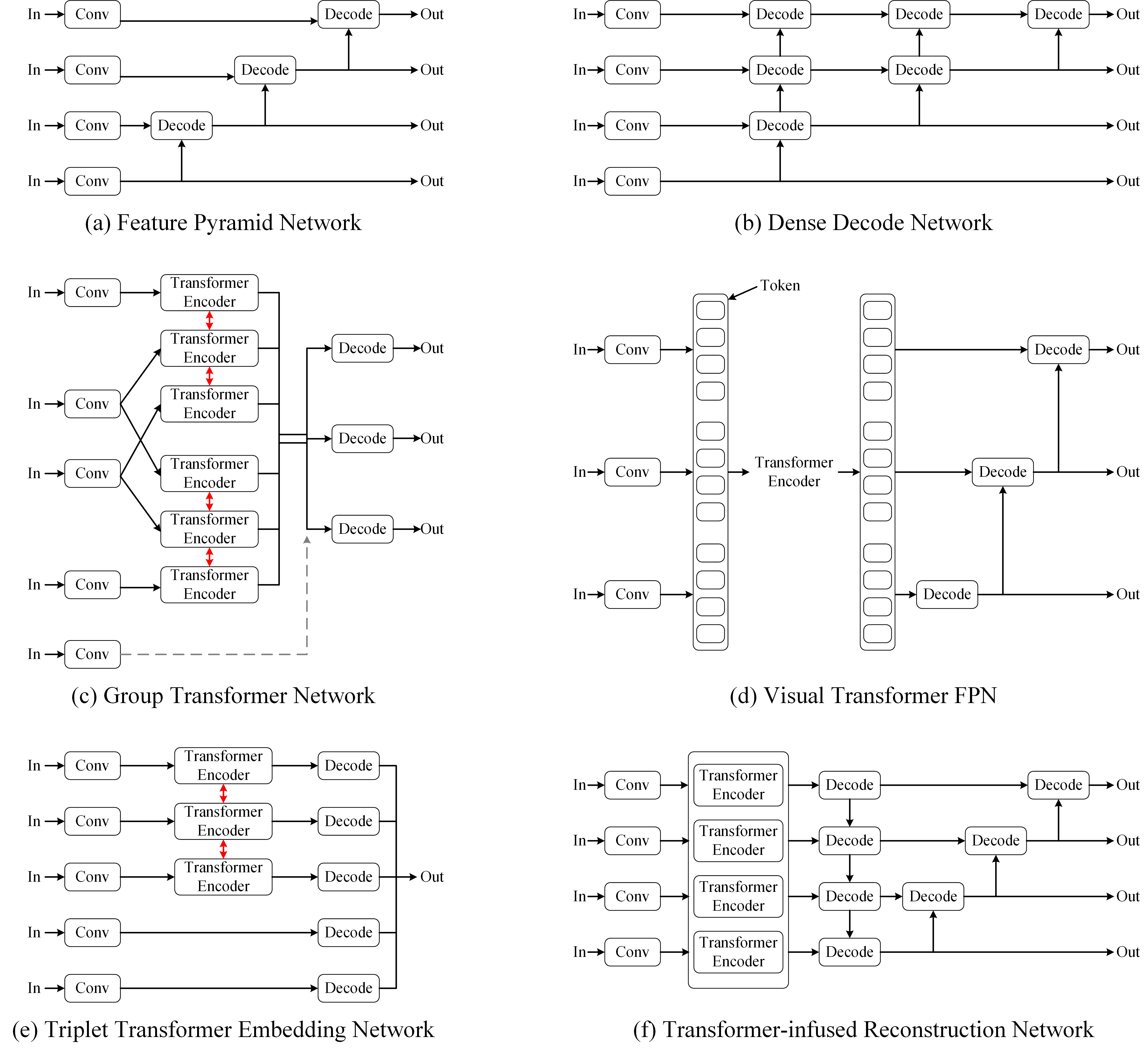}
	\caption{Comparison between (a) FPN framework, (b) dense decode network, (c) group transformer network, (d) visual transformer FPN, (e) triplet transformer embedding network, and our (f) transformer-infused reconstruction network.}
	\label{fig:architecture}
\end{figure}

Fig. \ref{fig:examples} shows a part of comparisons between our model and other representative methods, including CATNet, HiDANet and TriTransNet. Our model delivers a better result for the RGB-D SOD task. Also, in Fig. \ref{fig:architecture}, we illustrate the framework difference between our model and the feature pyramid network (FPN) \cite{FPN}, dense decoding network \cite{dense}, group transformer network (GroupTransNet) \cite{fang2024grouptransnet}, visual transformer FPN \cite{wu2020visual}, and triplet transformer embedding network (TriTransNet) \cite{liu2021tritransnet}. The traditional models feature pyramid network and dense decoding network, only leverage the simplest decoder without transformer, leading to the worst performance. Our proposed model, however, divides the transformer into four stages, each of which can be fed into the decoder separately and independently. Hence, such parallelization in our model can speed up the whole learning process. On the contrary, the recent advanced models that leverage transformer, such as TriTransNet and GroupTransNet, are designed to repeatedly execute transformer multiple times, leading to high computation costs. Moreover, we have no shared parameters between stages. This helps us to explore the discriminative features for each stage. Therefore, compared with TriTransNet and GroupTransNet, we leverage fewer parameters, consume less cost, and gain better performance, which significantly distinguishes our work from previous studies. Furthermore, extensive experiments are conducted on six benchmark datasets, and the results indicate that the proposed model achieves competitive performance against 24 RGB-D SOD methods.

In summary, the main contributions of this paper can be described as follows:
\begin{itemize}
	\item We present a dual mutual learning network with global-local awareness (GL-DMNet) for the RGB-D SOD, which constructs a joint transformer-based and CNN-based network to extract features from RGB images and depth inputs.
	\item To fully exploit the global-local dependencies between two modalities, we design a position mutual fusion (PMF) module and a channel mutual fusion (CMF) module for cross-modality fusion.
	\item We also develop a cascade transformer-infused reconstruction (CTR) decoder to augment the global-local awareness of multi-level fusion features.
	\item The proposed method is evaluated on six publicly available datasets under four widely used metrics. Compared with 24 state-of-the-art approaches, GL-DMNet achieves superior performance. 
\end{itemize}

\section{Related work}
\subsection{RGB-D salient object detection}
Early RGB-D SOD methods usually design handcrafted features and various fusion strategies to integrate RGB and depth cues. Following this direction, numerous models have been proposed to detect salient objects \cite{ren2015exploiting,feng2016local,song2017depth}. However, they merely regard the depth stream as auxiliary information, which results in unsatisfactory performance.


With the rapid development of deep learning, CNN-based methods have achieved remarkable progress \cite{SIP_D3Net2021TNNLS, yi2022cross}. 
Wang et al. \cite{wang2023dcmnet} introduce a depth decomposition and recomposition module to filter out low-quality depth maps and enhance the quality of detrimental depth maps.
Zhang et al. \cite{zhang2024feature} present a two-stage model, including an image generation stage that produces pseudo-depth images from RGB inputs and a saliency reasoning stage that calibrates original depth images with pseudo-depth images and performs cross-modal feature fusion.
Zeng et al. \cite{zeng2024airsod} introduce a lightweight RGB-D saliency method named AirSOD, which incorporates a parameter-free parallel attention-shift convolution, multi-level multi-modal feature fusion, and multi-path enhancement to achieve a favorable balance between efficiency and performance.
Xiao et al. \cite{xiao2024dgfnet} construct a depth-guided fusion module to enhance the fusion of RGB and depth features, emphasizing attention and weighting mechanisms to augment spatial representations in saliency regions and effectively reducing the gap between modalities.
Cheng et al. \cite{cheng2023depth} propose a depth-induced gap-reducing network that utilizes cross-modality interaction blocks and interference degree mechanisms in two branches for effective cross-modality feature fusion and refinement while mitigating unexpected noise. 
Sun et al. \cite{CATNet} designed a Cascaded and Aggregated Transformer Network (CATNet) to establish connections between features of different scales to prevent information loss and redundancy.
Li et al. \cite{li2024robust} employed a dynamic searching process module and a dual-branch consistency learning mechanism to address the weak richness of pixel training samples (WRPS) and the poor structural integrity of salient objects (PSIO) and proposed an edge-region structure-refinement loss for precise segmentation. 
Zhong et al. \cite{zhong2024magnet} utilized a Multi-scale Awareness Fusion Module (MAFM) for low-level features and a Global Fusion Module (GFM) for high-level features, thereby enriching low-level information while performing a global analysis of multi-modal semantic information.

Despite the success existing methods have achieved, they still face some limitations which hinder further improvement in the area of RGB-D SOD. First, most of them focus on handling low-quality images to strengthen the information expressiveness and thus mitigate the noise caused by low quality. However, they ignore the inherent discrepancy between different modalities (i.e., RGB and depth modalities), which cannot be solved by only improving the quality of images. Second, there is a natural trade-off between the RGB and depth information. Sometimes depth delivers more significance to the SOD task than RGB information, while sometimes RGB information could play a more leading role \cite{zhang2019synthesizing}. Therefore, our scenario's core research question is how to efficiently utilize them together and achieve a double-win balance. Moreover, the features learned during each stage affect global learning differently. Unfortunately, most existing methods integrate them without identifying their global awareness, making system performance suboptimal. The mentioned issues encourage our work in this paper.

On the one hand, we aim to model the inherent discrepancy across modalities by leveraging PMF and CMF. We seek to learn the relevance between modalities and achieve better trade-offs. On the other hand,  we aim to explore the local-global awareness of each stage's learned features to ensure their accurate contributions to the SOD prediction. We hope our work could deliver an improvement to existing methods for the RGB-D SOD task and provide some innovations for other researchers in this area.

\subsection{Vision Transformer}


Transformer is initially proposed to address the limitations of recurrent and convolutional neural networks in handling sequential data \cite{vaswani2017attention}. The unique ability to handle long-range dependencies and high parallelization make it a versatile architecture and has been applied to various domains within the field of computer science \cite{dosovitskiy2020image,wang2021pyramid,liu2021swin}. The transformer-based models also open up new directions in RGB-D SOD.
Liu et al. \cite{liu2021visual} design a pure transformer framework that utilizes the T2T-ViT to divide images into patches and the RT2T transformation to decode patch tokens to saliency maps. 
Liu et al. \cite{SwinNet} propose SwimNet to explore the advantages of CNN and transformer in modeling local and global features. 
Cong et al. \cite{cong2023point} propose a cross-modality point-aware interaction module to simplify the process by grouping corresponding point features from different modalities, thereby addressing the cross-modality feature interaction challenge in transformer-based models.
Wang et al. \cite{wang2023uniform} construct a novel and unified transformer-based structure to globally fuse multi-scale multi-modal features and enhance features from the same modality.
Gao et al. \cite{gao2024tsvt} propose a novel token sparsification transformer framework, dynamically sparsifying tokens and extracting global-local multi-modality features based on the most informative tokens.
Liu et al. \cite{liu2021tritransnet} introduce a triplet transformer scheme called TritransNet that shares parameters among different stages, to model the long-range correlation between them globally. TritransNet servers as a supplementary encoding strategy for feature fusion, which also inspires our work. Nonetheless, the utilization of sharing parameters cannot capture the discriminate features of each stage. Meanwhile, executing the large parameters from the global view would be time-consuming for each stage. This motivates us to design an independent encoder for each stage while leveraging a global-aware strategy to protect the correlation among stages. 
Wu et al. \cite{wu2024transformer} propose a transformer-based fusion method with pixel-level contrastive learning to explore the potential correlation between the inter and intra-pixel interactions over modalities. Its core lies in the attention mechanism for cross-modality fusion. However, the large number of its CIPT modules leads to a severe overload of fusion strategies, which affects the efficiency of the whole model learning.

Encouraged by the promising results from TriTransNet \cite{liu2021tritransnet}, we attempted to further enhance the feature representation ability of multi-modal integrated features by employing the transformer-based architecture, aiming to achieve the interaction and fusion of long-distance information across stages.

\subsection{Attention mechanism}
In various vision tasks, attention mechanisms are fully exploited to focus on the pivotal information in the image, and have been proven to be an indispensable means for performance improvement, such as convolutional block \cite{woo2018cbam}, selective kernel block \cite{li2019selective}, strip pooling block \cite{hou2020strip}, and coordinate attentions \cite{hou2021coordinate}.

When being applied to RGB-D SOD, the above attention mechanisms enable the encoder to flexibly discern subtle differences between modalities, which provides an opportunity for further exploration. Also, they aim to deepen the integration of distinct modal features and exploit complementary information more effectively.
Liu et al. \cite{liu2022learning} leverage cross-modal attention propagation, which contains a contrast inference and incorporates a selective attention mechanism to address the challenges posed by low-quality depth data.
Feng et al. \cite{feng2022encoder} propose a cross-modal mutual guidance module to model the interdependencies between channels to implement the global guidance and local refinement of the salient region. 
Wang et al. \cite{wang2022boosting} introduce a depth integration module that merges three feature maps by leveraging depth-aware and depth-dispelled features to enhance the complementary information from RGB and depth modalities.
Cong et al. \cite{cong2022cir} propose a smAR unit to reduce feature redundancy and emphasize important spatial locations, and also a cmWR unit to refine multi-modality features by considering cross-modality complementary information and global contextual dependencies.
Fang et al. \cite{fang2023m2rnet} employ the adjacent interactive aggregation module to leverage the parallel interaction of progressive and jumping connections to gradually learn information in abundant resolution, enhancing the aggregation of high-level, middle-level, and low-level features.
Zhang et al. \cite{zhang2023C2DFNet} utilize a modal-specific dynamic enhanced module to adaptively enhance intra-modality features and a scene-aware dynamic fusion module to achieve dynamic feature selection between RGB and depth modalities.
Wu et al. \cite{wu2023hidanet} design a granularity-based attention module to strengthen saliency detection by generating distinct local-spatial attentional regions through depth granularity and improve feature discriminatory power by applying local channel attention.

Nevertheless, in our work, attention mechanisms are utilized for feature enhancement and the interaction and fusion of cross-modality features. Inspired by the dual attention network \cite{fu2019dual}, we design the PMF and CMF modules to model semantic interdependencies both spatial and channel dimensions, promoting interactions between cross-modality features.

\begin{figure*}[!t]
\centering
\includegraphics[width=1.00\textwidth]{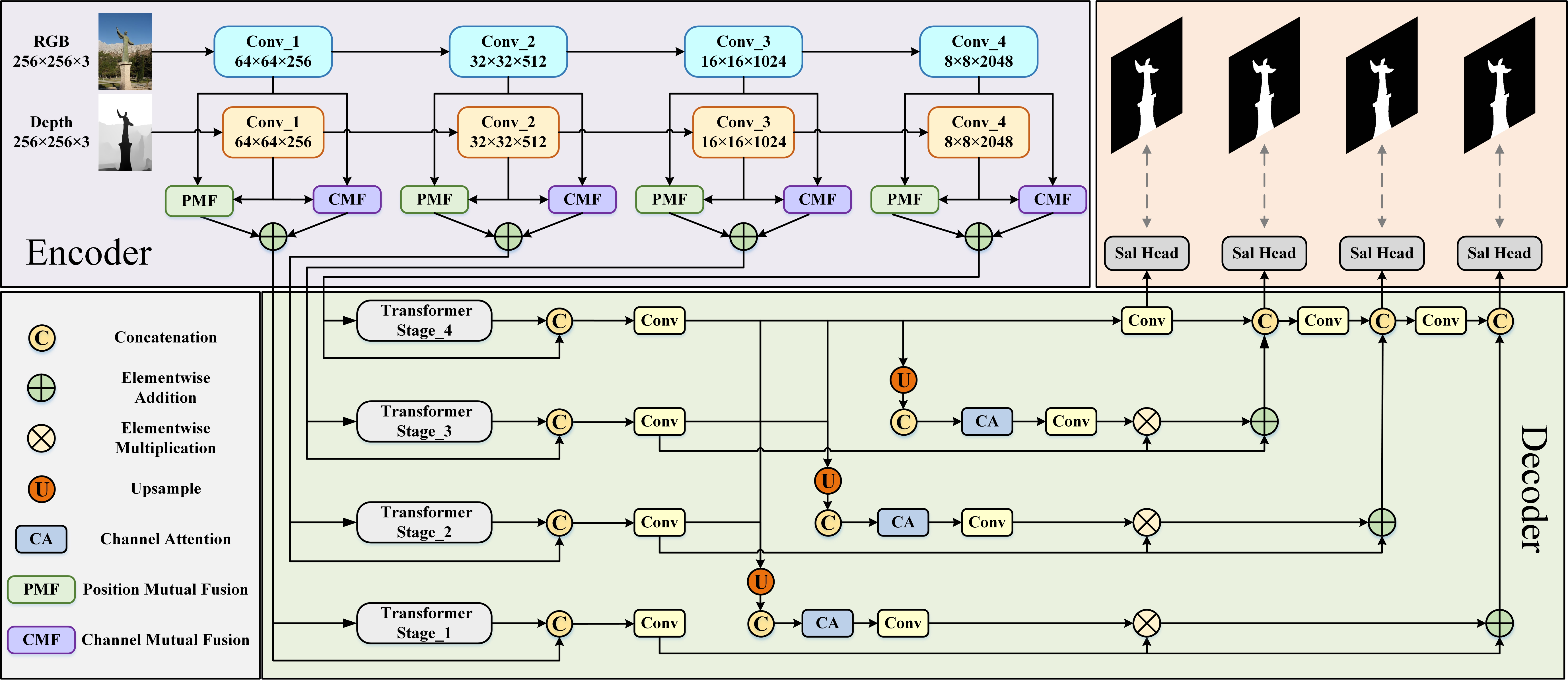}
\caption{Detailed framework of the proposed GL-DMNet. We adopt the ResNet-50 network to extract features of RGB and depth inputs, respectively. Then, position mutual fusion (PMF) and channel mutual fusion (CMF) are proposed to fuse the multi-modal features. The fused features of all stages are decoded by the cascade transformer-infused reconstruction network. The saliency head \cite{fan2021rethinking} is also added to generate the final predicted feature maps.}
\label{fig:GL-DMNet}
\end{figure*}

\section{Methodology}
\subsection{Overview}
The overall framework of the proposed GL-DMNet is shown in Fig. \ref{fig:GL-DMNet}, which consists of a multi-modal feature encoder, dual mutual fusion module, and cascade transformer-infused reconstruction decoder. 

Inspired by the triplet-transformer embedding architecture \cite{liu2021tritransnet} while pursuing a fair comparison, we first adopt the ResNet-50 \cite{ResNet} network to extract multi-level features from the original inputs of RGB image and depth map. They are denoted as $\mathcal{F}^{RGB}_{i}$ and $\mathcal{F}^{D}_{i}$, respectively, where $i \in \left \{1,2,3,4\right \} $ is the index of the stage in encoder backbones. To model semantic interdependencies in spatial and channel dimensions and further enhance the saliency feature representations, we feed the RGB and depth features into the proposed position mutual fusion module and channel mutual fusion module to generate the attentional cross-modality RGB-D features.

During the decoding phase, we devise a novel cascade transformer-infused reconstruction decoder, which decomposes the transformer network into four stages. The obtained fusion features of each stage are fed independently to establish long-range dependencies. Finally, we perform a progressive decoding reconstruction structure to derive the eventual saliency map. The descriptions of some symbols and notations are listed in Table. \ref{tab:Notations} to improve readability.

\begin{table}[!t]
	\centering
	\caption{Symbols and notations used in this paper and their descriptions.}
	\label{tab:Notations}
	\begin{tabular}{c|c}
		\hline
		Notations & Descriptions \\ \hline
		${\rm Conv}_{k}(\cdot)$ & $k \times k$ convolution layer together with BN and ReLU \\
  	${\rm MaxPool}(\cdot)$ & max-pooling operation \\
        ${\rm AvgPool}(\cdot)$ & average-pooling operation \\
		${\rm Cat}[\cdot;\cdot]$ & concatenation \\
        $\mathcal{M}(\cdot)$ & the moment normalization \\
        ${\mathcal{N}}(\cdot)$ & the L2 normalization \\
        ${\rm FC}(\cdot)$ & fully-connected layer \\
		${Trans_i}(\cdot)$ & the $i$-th stage of the transformer-based encoder \\
        ${\rm CA}(\cdot)$ & channel attention (CA) module  \\ 
		${\rm up}(\cdot)$ & upsampling with bilinear interpolations \\
		${\rm up}_{ori}(\cdot)$ & upsampling to the raw size of the input data\\
        $\odot$ & element-wise multiplication \\
   	$\otimes$ & matrix multiplication \\\hline
		$i$ & the $i$-th stage in the neural networks\\
		$\mathcal{F}_i$ & feature maps outside a module in the $i$-th stage \\
		$f_i$ & feature maps inside a module in the $i$-th stage \\\hline
	\end{tabular}
\end{table}

\subsection{Dual mutual learning module}
How to effectively integrate the cross-modality features has consistently been a challenging problem in RGB-D SOD tasks for a long time. Previous research primarily utilizes a combination of spatial and channel attention mechanisms to achieve cross-modal feature fusion. However, this approach often overlooks the distinctive information of each modality, resulting in a lack of complementary. Due to the inherent differences between RGB and depth modalities, where RGB images contain rich semantic information and depth maps provide spatial-geometric clues, aligning and bridging the potential gap between these modalities is crucial for accurate salient object detection \cite{zhang2021cross}.

\begin{figure*}[!t]
\centering
\includegraphics[width=1.00\textwidth]{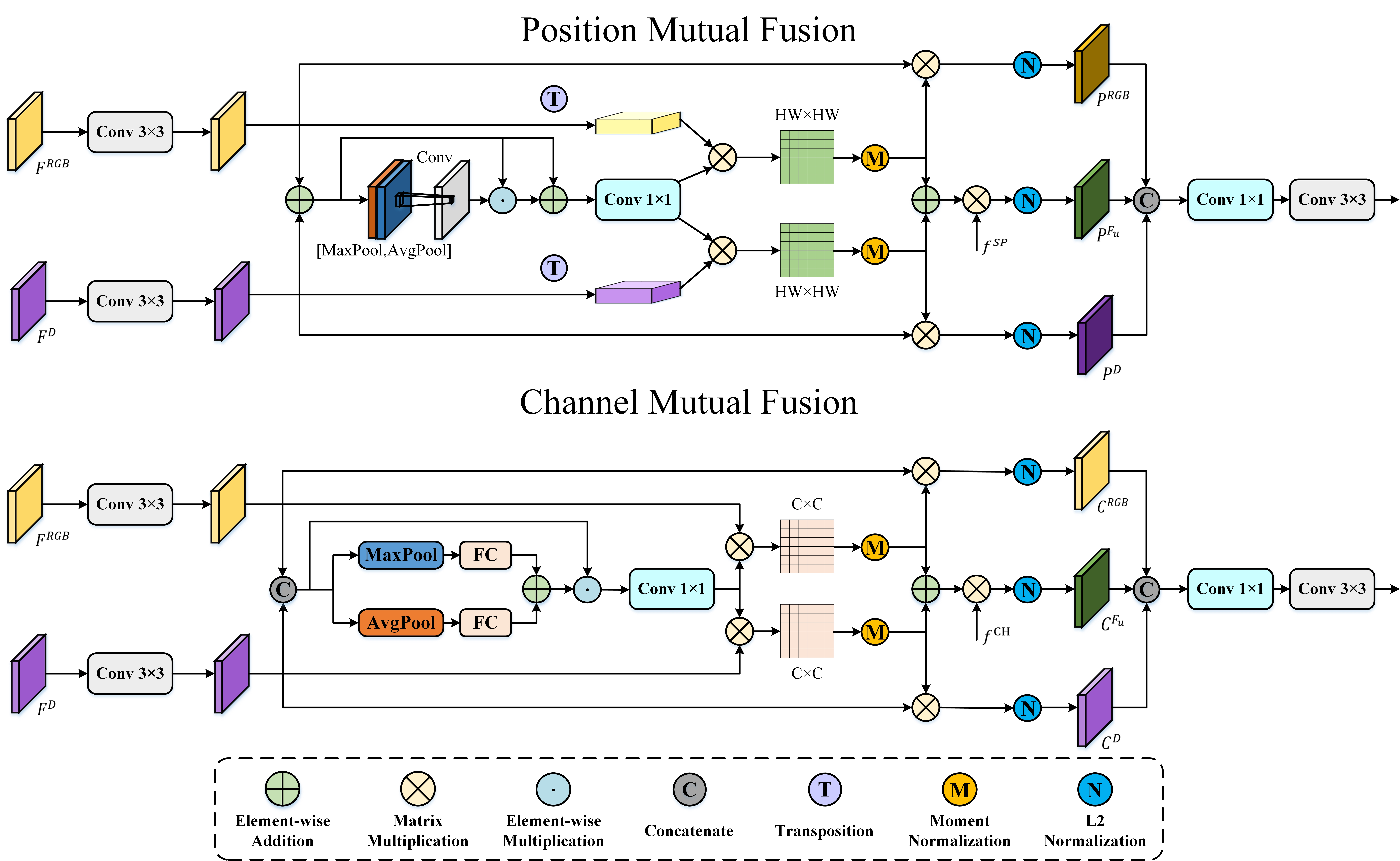}
\caption{The details of position mutual fusion (PMF) module and channel mutual fusion (CMF) module.}
\label{fig:PMF-CMF}
\end{figure*}

Inspired by the dual attention \cite{fu2019dual}, we propose a dual mutual learning module to enhance the coordination of multi-modal features, which models semantic interdependencies in spatial and channel dimensions. Even though most studies in SOD tasks capture the local features over modalities, they cannot globally explore the inherent relationship between different features (or objects). Our dual attention mechanism, however, allows the local features to learn from each other, globally propagating the information for them. It captures the rich global contextual information and could significantly improve the performance by modeling dependencies among local features. Specifically, as shown in Fig. \ref{fig:PMF-CMF}, the position mutual fusion module selectively aggregates features at each position by learning the spatial dependencies of the single-modal and fused features. Meanwhile, the channel mutual fusion module weights and updates each channel map by integrating the channel mapping relationship between single-modal and fused features.

Compared with previous work \cite{fang2023reliable,fang2022incremental}, we revise and improve the dual mutual learning to explore more complex cross-modality interaction information. Besides, we adopt some beneficial embedding-wise operations to improve the basic PMF and CMF. More importantly, our PMF and CMF allow the RGB and depth information to interact fully with their fusion result. A simple illustration in Fig. \ref{fig:Visualization} indicates how our PMF and CMF comprehensively explore fine-grained information from images.

\subsubsection{Position mutual fusion module}
To be specific, we first obtain the RGB features $\mathcal{F}^{RGB}_{i}$ and the depth features $\mathcal{F}^{D}_{i}$ from their corresponding backbones, and then apply two convolutional layers to reduce the number of channels:
\begin{align}
{f}^{RGB}_{i} &= Conv_{3}(Conv_{1}(\mathcal{F}^{RGB}_{i})), \\
{f}^{D}_{i} &= Conv_{3}(Conv_{1}(\mathcal{F}^{D}_{i})),
\end{align}
where ${\rm Conv}_{k}(\cdot)$ denotes a $k\times k$ convolutional layer together with a batch normalization (BN) layer \cite{ioffe2015batch} and a rectified linear unit (ReLU) activation function.

Subsequently, we perform element-wise addition to fuse the cross-modality features and adopt the spatial attention mechanism \cite{woo2018cbam} to acquire more valuable cross-modal fusion features. The entire process could be described as:
\begin{align}
{f}^{A}_{i} &= {f}^{RGB}_{i} + {f}^{D}_{i}, \\
{W}^{SP}_{i} &= Conv_{7}(Cat(MaxPool({f}^{A}_{i}), AvgPool({f}^{A}_{i}))), \\
{f}^{SP}_{i} &= Conv_{1}({f}^{A}_{i} \odot {W}^{SP}_{i} + {f}^{A}_{i}),
\end{align}
where $MaxPool(\cdot)$ and $AvgPool(\cdot)$ denote the max-pooling and average-pooling operations along channel dimension, respectively. In addition, ${\rm Cat}(\cdot,\cdot)$ and $\odot$ represent the concatenation operation and element-wise multiplication. 

Then, we reshape the original single-modal features (i.e., ${f}^{RGB}_{i}$ and ${f}^{D}_{i}$) and the fused features ${f}^{SP}_{i}$ to $\mathbb{R}^{C\times N}$, where $N=H\times W$ represents the number of pixels. The matrix multiplication operations are also employed to generate the spatial attention maps:
\begin{align}
{Ms}^{RGB}_{i} &= \mathcal{M}({f}^{SP}_{i} \otimes ({f}^{RGB}_{i})^T),\\
{Ms}^{D}_{i} &= \mathcal{M}({f}^{SP}_{i} \otimes ({f}^{D}_{i})^T),\\
{Ms}^{Fu}_{i} &= {Ms}^{RGB}_{i} + {Ms}^{D}_{i},
\end{align}
where $\otimes$ is the matrix multiplication, and $\mathcal{M}(x)=sign(x) \cdot x^{-1/2}$ is the moment normalization to ensure the stability and the appropriate scaling of the computed values.

Next, we refine these modality features by multiplying the positional weights and the original features. Also, the L2 normalization is utilized to ensure that the values in each channel belong to the valid probability distribution, further strengthening the stability of the network during training:
\begin{align}
{P}^{RGB}_{i} &= \mathcal{N}({f}^{RGB}_{i} \otimes {Ms}^{RGB}_{i}),\\
{P}^{D}_{i} &= \mathcal{N}({f}^{D}_{i} \otimes {Ms}^{D}_{i}),\\
{P}^{Fu}_{i} &= \mathcal{N}({f}^{SP}_{i} \otimes {Ms}^{Fu}_{i}),
\end{align}
where ${\mathcal{N}}(x)=x/\left \| x \right \| _{2}^{2} $ represents the L2 normalization.

\subsubsection{Channel mutual fusion module}
Similar to the position mutual fusion module, we employ concatenation operations to produce fused features, following the channel attention mechanism \cite{woo2018cbam} to generate more discriminative features. This could be indicated as:
\begin{align}
{f}^{C}_{i} & = Cat({f}^{RGB}_{i}, {f}^{D}_{i}), \\
{W}^{CH}_{i} & = FC(MaxPool({f}^{C}_{i})) + FC(AvgPool({f}^{C}_{i})), \\
{f}^{CH}_{i} & = Conv_{1}({f}^{C}_{i} \odot {W}^{CH}_{i}),
\end{align}
where $FC(\cdot)$ composed of two fully-connected layers.

However, different from the position mutual fusion module, we directly adopt the matrix multiplication operations to calculate the channel attention maps:
\begin{align}
{Mc}^{RGB}_{i} &= \mathcal{M}({f}^{RGB}_{i} \otimes {f}^{CH}_{i}),\\
{Mc}^{D}_{i} &= \mathcal{M}({f}^{D}_{i} \otimes {f}^{CH}_{i}),\\
{Mc}^{Fu}_{i} &= {Mc}^{RGB}_{i} + {Mc}^{D}_{i}.
\end{align}

We also perform the matrix multiplications to generate the multi-modality features with long-range contextual representations:
\begin{align}
{C}^{RGB}_{i} &= \mathcal{N}({Mc}^{RGB}_{i} \otimes {f}^{RGB}_{i}),\\
{C}^{D}_{i} &= \mathcal{N}({Mc}^{D}_{i} \otimes {f}^{D}_{i}),\\
{C}^{Fu}_{i} &= \mathcal{N}({Mc}^{Fu}_{i} \otimes {f}^{CH}_{i}).
\end{align}

Finally, we aggregate the outputs of the two mutual fusion modules and use an element-wise summation to complete the eventual feature fusion, which will absorb more semantic information. This process could be formulated as:
\begin{align}
{f}^{PMF}_{i} &= Conv_3(Conv_1(Cat({P}^{RGB}_{i}, {P}^{D}_{i}, {P}^{Fu}_{i}))),\\
{f}^{CMF}_{i} &= Conv_3(Conv_1(Cat({C}^{RGB}_{i}, {C}^{D}_{i}, {C}^{Fu}_{i}))),\\
\mathcal{F}^{Fus}_{i} &= {f}^{PMF}_{i} + {f}^{CMF}_{i}
\end{align}

In conclusion, the elaborately designed dual mutual learning component fully learns the potential features while highlighting each modality's core cues. It considers the correlation and complementarity between different modalities, thereby establishing long-range contextual dependence across modalities.

\subsection{Cascade Transformer-Infused Reconstruction Decoder}
Despite the advantages of CNNs, their limited receptive field severely restricts the capability to capture global features. Therefore, we integrate transformers to augment the global-local awareness of multi-level fusion features, enabling the proposed method to inherit the strengths of both CNN and Transformer. The decoded feature maps at each stage contain more spatial details and could contribute uniquely to the final prediction.

\textbf{Transformer Embedding:} We employ PVTv2-B2 \cite{wang2022pvt} to learn global representations of the obtained multi-level fusion features ${F}^{u}_i$, because it could simultaneously protect local continuity by overlapping patch embedding. PVTv2 is divided into four stages to generate features of different scales. At each stage, the input image is separately decomposed into patches by a trainable linear projection layer. Then, these patches are fed into a transformer-based encoder with several layers. Afterwards, the output is reshaped into feature maps for multi-level prediction tasks. This design could be described as:
\begin{equation}
{f}^{t}_i = Trans_i \left ( \mathcal{F}^{Fus}_{i} \right ),
\end{equation}
where $Trans_i(\cdot)$ indicates the $i$-th stage of the transformer-based encoder. Moreover, another advantage of PVTv2 is that the resolution of output feature maps of each stage remains the same as input features, making it an easy plug-and-play component to integrate with the proposed architecture.

\textbf{Multi-level Feature Reconstruction:} Naturally, there are differences between features at different levels. Thus, directly aggregating them may lead to information loss or redundancy \cite{CATNet}. In our proposed decoder scheme, we utilize high-level features to guide the decoding of low-level features while filtering the inherent differences among features. More importantly, our method focuses on the complementary and consistent information between adjacent levels.

Specifically, we first cascade the outputs of the transformer encoder with original fusion features and then adopt two successive convolutions to reduce the number of channels, which could be formulated as:
\begin{equation}
{f}^{t}_i = Conv_{3}(Conv_{1}(Cat({f}^{t}_i, \mathcal{F}^{Fus}_{i})).
\end{equation}

Subsequently, we connect the higher-level features through the dense upsampling and concatenation operations. The channel attention (CA) \cite{hu2018squeeze} mechanism is also employed to model the significant correlation among different channels, which could be expressed as:
\begin{equation}
{f}^{att}_i = Conv_{3}(Conv_{1}(CA(Cat[up({f}^{t}_{i+1}),..., up({f}^{t}_4)]))),
\end{equation}
where $up(\cdot)$ denotes the upsampling operation via the bilinear interpolation that reshapes ${f}^{t}_{i+1}$ to same resolution with ${f}^{t}_{i}$. 

Also, we utilize an element-wise multiplication to integrate multi-level fusion features and leverage a residual connection to preserve the original features:
\begin{equation}
{f}^{res}_i = {f}^{att}_i\odot {f}^{t}_i + {f}^{t}_i.
\end{equation}

Finally, we sequentially concatenate features from adjacent stages, progressively generating accurate saliency maps:
\begin{align}
{f}^{out}_4 &= Conv_{3}(Conv_{1}({f}^{res}_{4})),\\
{f}^{out}_i &= Conv_{3}(Conv_{1}(Cat({f}^{res}_{i}, {f}^{out}_{i+1}))), i=1,2,3.
\end{align}

In addition, we add deep supervisions to the outputs ${f}^{out}_i$ of the decoder to speed up convergence. The predicted saliency maps could be formulated as,
\begin{equation}
S_i = up_{ori}(Conv_{1}(Conv_{3}({f}^{out}_i)))
\end{equation}
where $S_i$ represents the predictions in the $i$-th stage. ${up}_{ori}(\cdot)$ upsamples the feature maps to the original resolution of the input image. Only $S_1$ is used as the final saliency map, while the other three predictions are omitted in the test phase.

\subsection{Loss function}
To obtain high-quality saliency maps with clear boundaries, we decide to employ the BCE loss $\ell_{bce}$ \cite{BCE} and the $\ell_{iou}$ \cite{IOU} to train our proposed model.

Binary Cross-Entropy (BCE) loss is the most widely-used loss in segmentation tasks, which computes the discrepancy between predicted saliency maps and ground truth binary labels. It weights both foreground and background pixels equally and accurately distinguishes salient and non-salient regions. It could be defined as below: 
\begin{equation}
\ell_{bce} = - \sum\limits_{w,h=1}^{W,H} {[{G_{wh}}\log({S_{wh}}) + (1 - {G_{wh}})\log(1 - {S_{wh}})]}
\end{equation}

Intersection over Union (IoU) loss, however, measures the overlap between predicted saliency maps and ground truth, guiding the model to focus more on the foreground and produce more accurate spatially-aligned predictions. It could be defined as:
\begin{equation}
\ell_{iou} = 1 -  {\frac{\sum\limits_{w,h=1}^{W,H} {S_{wh}}{G_{wh}}} {\sum\limits_{w,h=1}^{W,H} ({S_{wh}} + {G_{wh}} - {S_{wh}}{G_{wh}} )}}
\end{equation}

We adopt a multi-task learning strategy to comprehensively learn the hierarchical saliency information from multi-level outputs by jointly training the model with the multiple-auxiliary side outputs. Hence, the total loss of our training could be indicated as:
\begin{equation}
\ell_{total}= \sum\limits_{i=1}^{4} {\lambda_i ({\ell_i^{bce}({S_i},G)} + \ell_i^{iou}({S_i},G)}),
\end{equation}
where $G$ denotes the ground truth, and $S_i$ refers to the saliency map predicted from the $i-th$ level of the decoder, which is upsampled to form the same resolution mask as the ground truth. $\lambda_i$ is the weight of the $i-th$ level. The weight set $\lambda$ follows $\left \{ 0.8,0.6,0.4,0.2 \right \}$ to correlate with the network hierarchies.

\begin{table*}[!t]
    \centering
    \caption{Quantitative results compared with state-of-the-art RGB-D SOD methods. ``-'' means that the results are unavailable since the authors did not release them. $\uparrow$ ($\downarrow$) indicates the larger (smaller), the better. The best and the second best results are highlighted in bold and underline, respectively.}
    \label{tab:quantitative}
\resizebox{\linewidth}{!}{
\begin{tabular}{c|cccc|cccc|cccc}
\hline
\multirow{2}{*}{Model} & \multicolumn{4}{c|}{SIP \cite{SIP_D3Net2021TNNLS}} & \multicolumn{4}{c|}{DUT-RGBD \cite{DMRA-DUT}} & \multicolumn{4}{c}{NJUD \cite{NJUD}} \\ \cline{2-13} 
 & $E_\xi$ $\uparrow$ & $S_\alpha$ $\uparrow$ & $F_\beta$ $\uparrow$ & MAE $\downarrow$ & $E_\xi$ $\uparrow$ & $S_\alpha$ $\uparrow$ & $F_\beta$ $\uparrow$ & MAE $\downarrow$ & $E_\xi$ $\uparrow$ & $S_\alpha$ $\uparrow$ & $F_\beta$ $\uparrow$ & MAE $\downarrow$ \\ \hline
Ours & \textbf{0.936} & \textbf{0.896} & \textbf{0.915} & \textbf{0.041} & \textbf{0.962} & \textbf{0.931} & \textbf{0.947} & \textbf{0.029} & {\ul 0.950} & \textbf{0.920} & \textbf{0.930} & \textbf{0.033} \\ \hline
FCFNet(24TCSVT) & - & - & - & - & {\ul 0.956} & 0.924 & 0.931 & 0.032 & \textbf{0.953} & {\ul 0.918} & 0.923 & {\ul 0.034} \\
AirSOD(24TCSVT) & 0.903 & 0.858 & 0.887 & 0.060 & 0.946 & 0.891 & 0.920 & 0.048 & 0.944 & 0.908 & 0.918 & 0.039 \\
MIRV(24TCSVT) & 0.927 & 0.877 & 0.894 & {\ul 0.049} & - & - & - & - & 0.929 & 0.890 & 0.899 & 0.046 \\
DHFR(23TIP) & 0.902 & 0.843 & 0.874 & 0.064 & - & - & - & - & 0.936 & 0.893 & 0.901 & 0.040 \\
S3Net(23TMM) & {\ul 0.933} & 0.875 & 0.891 & 0.051 & 0.939 & 0.912 & 0.922 & 0.035 & 0.944 & 0.913 & 0.928 & {\ul 0.034} \\
M2RNet(23PR) & 0.921 & 0.882 & 0.902 & {\ul 0.049} & 0.935 & 0.903 & 0.925 & 0.042 & 0.904 & 0.910 & 0.922 & 0.049 \\
HINet(23PR) & 0.899 & 0.856 & 0.880 & 0.066 & {\ul -} & - & - & - & 0.945 & 0.915 & 0.914 & 0.039 \\
DLMNet(22MM) & 0.855 & 0.777 & 0.789 & 0.109 & 0.897 & 0.846 & 0.865 & 0.074 & 0.859 & 0.808 & 0.818 & 0.097 \\
MobileSal(22TPAMI) & 0.916 & 0.873 & 0.898 & 0.053 & 0.950 & 0.896 & 0.912 & 0.041 & 0.942 & 0.905 & 0.914 & 0.041 \\
JLDCF(22TPAMI) & 0.923 & 0.881 & {\ul 0.905} & 0.050 & 0.938 & 0.905 & 0.924 & 0.043 & 0.935 & 0.902 & 0.912 & 0.041 \\
DENet(22TIP) & 0.908 & 0.852 & 0.873 & 0.061 & - & - & - & - & 0.920 & 0.882 & 0.893 & 0.050 \\
CCAFNet(22TMM) & 0.915 & 0.877 & 0.881 & 0.054 & 0.941 & 0.905 & 0.915 & 0.036 & 0.920 & 0.910 & 0.911 & 0.037 \\
DWD(22TMM) & - & - & - & - & 0.902 & 0.864 & 0.853 & 0.072 & 0.927 & 0.886 & 0.876 & 0.050 \\
MoADNet(22TCSVT) & 0.911 & 0.865 & 0.890 & 0.058 & 0.945 & 0.907 & 0.920 & 0.033 & 0.929 & 0.901 & 0.907 & 0.042 \\
MMNet(22TCSVT) & 0.871 & 0.824 & 0.860 & 0.080 & 0.951 & 0.920 & 0.939 & 0.032 & 0.922 & 0.910 & 0.918 & 0.038 \\
EMANet(22PR) & - & - & - & - & 0.951 & 0.920 & 0.937 & 0.032 & 0.946 & 0.914 & 0.923 & 0.035 \\
DCF(21CVPR) & 0.920 & 0.873 & 0.899 & 0.052 & {\ul 0.956} & 0.924 & 0.940 & 0.031 & 0.940 & 0.903 & 0.917 & 0.039 \\
CDINet(21MM) & 0.911 & 0.875 & 0.903 & 0.055 & {\ul 0.956} & {\ul 0.926} & {\ul 0.944} & {\ul 0.030} & 0.944 & {\ul 0.918} & 0.827 & 0.036 \\
DFM-Net(21MM) & 0.926 & {\ul 0.883} & 0.887 & 0.051 & 0.945 & 0.913 & 0.928 & 0.039 & 0.947 & 0.906 & 0.910 & 0.042 \\
BiANet(21TIP) & 0.920 & {\ul 0.883} & 0.904 & 0.053 & - & - & - & - & 0.928 & 0.915 & {\ul 0.929} & 0.039 \\
DQSD(21TIP) & 0.900 & 0.863 & 0.890 & 0.065 & 0.889 & 0.844 & 0.859 & 0.073 & 0.912 & 0.898 & 0.910 & 0.051 \\
DRLF(21TIP) & 0.891 & 0.850 & 0.868 & 0.071 & 0.870 & 0.825 & 0.851 & 0.080 & 0.901 & 0.886 & 0.883 & 0.055 \\
cmSalGAN(21TMM) & 0.904 & 0.864 & 0.889 & 0.064 & 0.904 & 0.867 & 0.887 & 0.067 & 0.923 & 0.903 & 0.910 & 0.047 \\
IRFRNet(21TNNLS) & 0.921 & 0.879 & 0.881 & 0.054 & 0.951 & 0.919 & 0.924 & 0.035 & 0.945 & 0.909 & 0.908 & 0.040 \\ \hline
\hline
\multirow{2}{*}{Model} & \multicolumn{4}{c|}{STEREO \cite{STEREO}} & \multicolumn{4}{c|}{NLPR \cite{NLPR}} & \multicolumn{4}{c}{SSD \cite{SSD}} \\ \cline{2-13} 
 & $E_\xi$ $\uparrow$ & $S_\alpha$ $\uparrow$ & $F_\beta$ $\uparrow$ & MAE $\downarrow$ & $E_\xi$ $\uparrow$ & $S_\alpha$ $\uparrow$ & $F_\beta$ $\uparrow$ & MAE $\downarrow$ & $E_\xi$ $\uparrow$ & $S_\alpha$ $\uparrow$ & $F_\beta$ $\uparrow$ & MAE $\downarrow$ \\ \hline
Ours & \textbf{0.947} & {\ul 0.908} & \textbf{0.918} & \textbf{0.037} & {\ul 0.962} & \textbf{0.927} & \textbf{0.926} & {\ul 0.022} & \textbf{0.920} & \textbf{0.874} & \textbf{0.887} & \textbf{0.045} \\ \hline
FCFNet(24TCSVT) & \textbf{0.947} & 0.906 & 0.906 & {\ul 0.038} & 0.960 & 0.924 & 0.911 & 0.024 & - & - & - & - \\
AirSOD(24TCSVT) & 0.939 & 0.895 & 0.900 & 0.043 & \textbf{0.963} & 0.924 & {\ul 0.923} & 0.023 & - & - & - & - \\
MIRV(24TCSVT) & 0.937 & 0.891 & 0.900 & 0.042 & 0.954 & 0.914 & 0.914 & 0.025 & - & - & - & - \\
DHFR(23TIP) & 0.935 & 0.884 & 0.896 & 0.043 & 0.950 & 0.904 & 0.901 & 0.027 & 0.911 & 0.858 & 0.869 & 0.051 \\
S3Net(23TMM) & {\ul 0.945} & \textbf{0.913} & \textbf{0.918} & {\ul 0.038} & {\ul 0.962} & \textbf{0.927} & {\ul 0.923} & \textbf{0.021} & - & - & - & - \\
M2RNet(23PR) & 0.929 & 0.899 & 0.913 & 0.042 & 0.941 & 0.918 & 0.921 & 0.033 & - & - & - & - \\
HINet(23PR) & 0.933 & 0.892 & 0.883 & 0.049 & 0.957 & 0.922 & 0.906 & 0.026 & 0.916 & 0.865 & 0.852 & 0.049 \\
DLMNet(22MM) & 0.886 & 0.833 & 0.838 & 0.079 & 0.815 & 0.795 & 0.768 & 0.081 & 0.841 & 0.788 & 0.796 & 0.101 \\
MobileSal(22TPAMI) & 0.940 & 0.903 & 0.906 & 0.041 & 0.961 & 0.920 & 0.916 & 0.025 & 0.914 & 0.862 & 0.863 & 0.052 \\
JLDCF(22TPAMI) & 0.937 & 0.903 & {\ul 0.914} & 0.040 & 0.954 & {\ul 0.925} & \textbf{0.926} & 0.023 & - & - & - & - \\
DENet(22TIP) & 0.928 & 0.881 & 0.891 & 0.048 & 0.943 & 0.900 & 0.897 & 0.031 & 0.875 & 0.830 & 0.833 & 0.070 \\
CCAFNet(22TMM) & 0.921 & 0.891 & 0.887 & 0.044 & 0.952 & 0.922 & 0.909 & 0.026 & - & - & - & - \\
DWD(22TMM) & 0.933 & 0.899 & 0.887 & 0.046 & 0.936 & 0.906 & 0.882 & 0.038 & {\ul 0.917} & 0.861 & 0.832 & 0.049 \\
MoADNet(22TCSVT) & 0.931 & 0.896 & 0.901 & 0.043 & 0.950 & 0.918 & 0.908 & 0.024 & 0.900 & 0.854 & 0.863 & 0.057 \\
MMNet(22TCSVT) & 0.916 & 0.884 & 0.896 & 0.046 & 0.955 & {\ul 0.925} & 0.919 & 0.023 & 0.912 & {\ul 0.871} & 0.872 & {\ul 0.047} \\
EMANet(22PR) & 0.939 & 0.901 & 0.911 & 0.040 & 0.955 & 0.924 & 0.922 & 0.024 & 0.909 & 0.870 & 0.875 & {\ul 0.047} \\
DCF(21CVPR) & 0.943 & 0.905 & {\ul 0.914} & \textbf{0.037} & 0.956 & 0.921 & 0.917 & 0.023 & 0.905 & 0.851 & 0.857 & 0.054 \\
CDINet(21MM) & - & 0.905 & 0.903 & 0.041 & 0.953 & \textbf{0.927} & {\ul 0.923} & 0.024 & 0.906 & 0.852 & 0.867 & 0.057 \\
DFM-Net(21MM) & 0.941 & 0.898 & 0.893 & 0.045 & 0.957 & 0.923 & 0.908 & 0.026 & - & - & - & - \\
BiANet(21TIP) & 0.929 & 0.903 & 0.910 & 0.044 & 0.955 & {\ul 0.925} & 0.921 & 0.025 & 0.901 & 0.867 & 0.870 & 0.051 \\
DQSD(21TIP) & 0.911 & 0.891 & 0.900 & 0.052 & 0.934 & 0.915 & 0.909 & 0.030 & 0.890 & 0.868 & {\ul 0.877} & 0.053 \\
DRLF(21TIP) & 0.915 & 0.888 & 0.878 & 0.050 & 0.935 & 0.902 & 0.904 & 0.032 & 0.879 & 0.834 & 0.859 & 0.066 \\
cmSalGAN(21TMM) & 0.932 & 0.900 & 0.910 & 0.050 & 0.948 & 0.922 & {\ul 0.923} & 0.027 & 0.851 & 0.791 & 0.764 & 0.086 \\
IRFRNet(21TNNLS) & 0.941 & 0.897 & 0.893 & 0.044 & 0.960 & 0.921 & 0.910 & 0.026 & 0.910 & 0.864 & 0.841 & 0.053 \\ \hline
\end{tabular}
}
\end{table*}

\section{Experiments}
\subsection{Datasets and evaluation metrics}
\textbf{Datasets:} We conduct experiments on six widely used datasets to validate our proposed GL-DMNet. \textbf{SIP} \cite{SIP_D3Net2021TNNLS} is a high-quality RGB-D dataset with 929 images that capture various human postures and movements. \textbf{DUT-RGBD} \cite{DMRA-DUT} consists of 800 indoor and 400 outdoor scene images paired with corresponding depth maps. \textbf{NJUD} \cite{NJUD} contains 2,003 stereo image pairs with diverse objects, complex scenarios, and ground-truth maps. \textbf{STEREO} \cite{STEREO}, which is the first collection of stereo images utilized for saliency analysis, includes 1,000 initial images and then retains 797 images after official updates. \textbf{NLPR} \cite{NLPR} includes 1,000 stereo images from 11 indoor and outdoor scene types. \textbf{SSD} \cite{SSD} is a small-scale dataset that only collects 80 natural images of left and right views from three stereoscopic movies. Following the most common setup of previous studies \cite{jin2022moadnet, wu2024transformer}, we use 700 images from NLPR, 800 pairs from DUT-RGBD, and 1,485 samples from NJUD as the training set. The remaining images with corresponding depth maps (and also the other three datasets) are used for testing. 

\textbf{Evaluation metrics:} We adopt five public evaluation metrics for the experiments, including E-measure ($E_\xi$) \cite{fan2018enhanced}, S-measure ($S_\alpha$) \cite{S-measure}, max F-measure ($F_\beta$) \cite{Achanta2009CVPR}, mean absolute error (MAE) \cite{Borji2015TIP}, and precision-recall (PR) curve. E-measure assesses the similarity by considering region-aware precision, recall, and harmonic mean. S-measure evaluates the structural similarity at both object and region levels. F-measure is the harmonic mean of precision and recall, which is more fairly evaluating the model's ability to capture salient regions. MAE measures the average absolute differences between predicted and ground-truth saliency maps. PR curve illustrates the trade-off balance between precision and recall from various threshold levels, which visually provides intuitive results when comparing the performance across different models. In summary, these selected metrics provide sufficient and comprehensive evaluations of structural similarity, region-aware performance, balanced precision-recall rates, and overall discrimination ability.

\subsection{Implementation details}
We conduct all experiments using PyTorch on an NVIDIA GeForce RTX 3090 GPU. The CNN-based network ResNet-50 \cite{ResNet} and transformer-based network PVTv2-B2 \cite{wang2022pvt} are employed to extract local and global features, respectively. During the training and test phases, we replicate the input depth map into three channels and resize all images to 256$\times$256 pixels. Additionally, different kinds of data augmentations, such as random cropping, flipping, rotation, and color enhancement, are employed during the training to prevent overfitting. We utilize the Adam optimizer \cite{kingma2014adam} to train the proposed network in an end-to-end manner with a batch size of 4 for 200 epochs. During the initial 30 epochs, we freeze the ResNet-50 networks to train the other parameters of the model. For the subsequent 30 epochs, we unfreeze the ResNet-50 network and freeze the Transformer network. After 60 epochs, we unfreeze all networks to fine-tune the parameters of our model. The initial learning rate is set to 1e-4 with a decay factor of 0.97. Last, the entire training process takes approximately 7 hours to converge. The inference time for images with 256$\times$256 size is 40 frames per second (FPS).

\begin{figure*}[!t]
	\centering
	\includegraphics[width=1.00\textwidth]{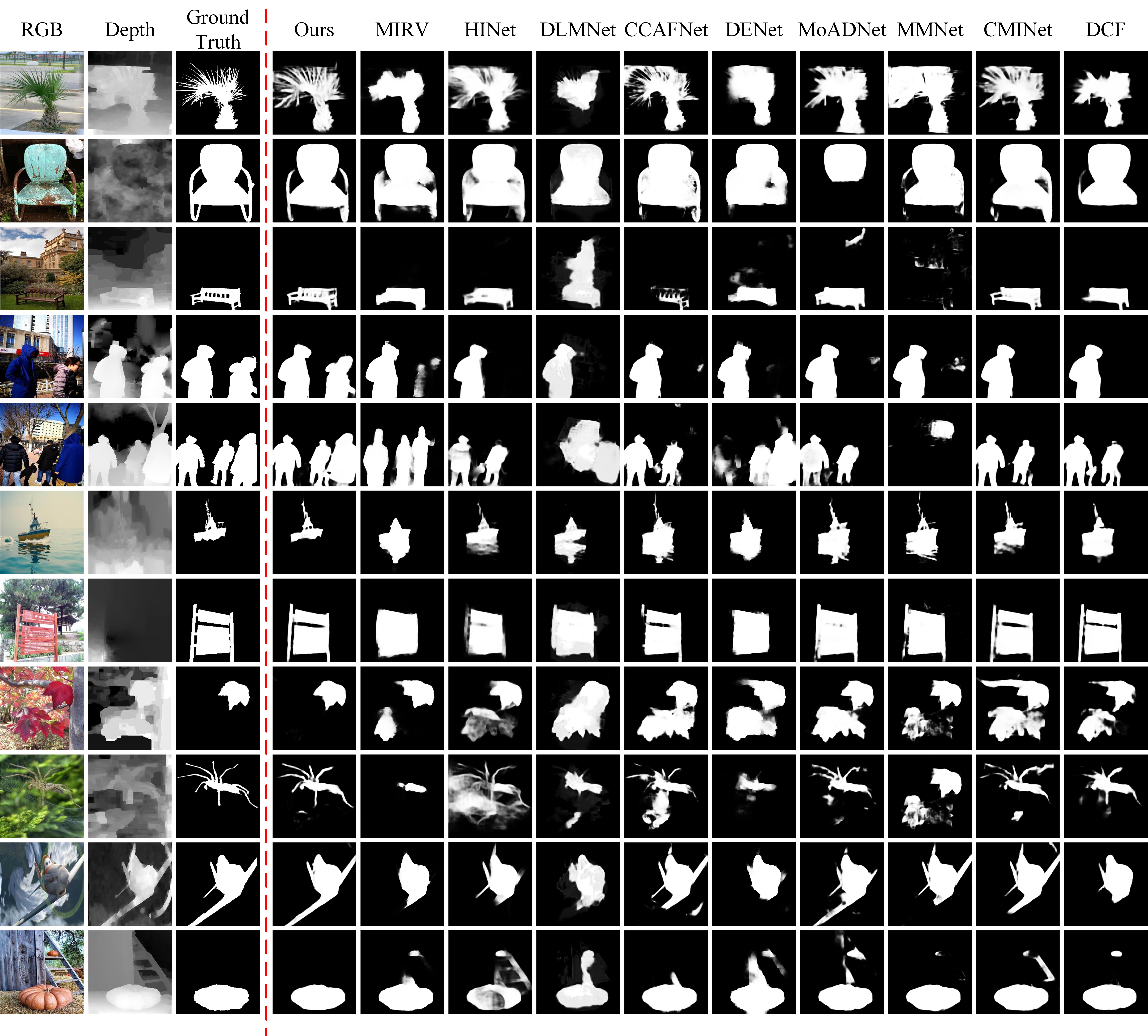}
	\centering
	\caption{Visual comparisons of the proposed GL-DMNet and other state-of-the-art RGB-D SOD methods, including MIRV \cite{li2024mutual}, HINet \cite{bi2023cross}, DLMNet \cite{yang2022depth}, CCAFNet \cite{zhou2022ccafnet}, DENet \cite{xu2022weakly}, MoADNet \cite{jin2022moadnet}, MMNet \cite{gao2022unified}, CMINet \cite{yi2022cross} and DCF \cite{sun2021deep}. Our approach obtains competitive performance in a variety of challenging scenarios.}
	\label{fig:qualitative}
\end{figure*}

\subsection{Comparisons with state-of-the-art methods}
We compare the proposed GL-DMNet with 24 state-of-the-art RGB-D SOD models, including FCFNet \cite{zhang2024feature}, AirSOD \cite{zeng2024airsod}, MIRV \cite{li2024mutual}, DHFR \cite{liu2023deep}, S3Net \cite{zhu2023s}, M2RNet \cite{fang2023m2rnet}, HINet \cite{bi2023cross}, DLMNet \cite{yang2022depth}, MobileSal \cite{wu2022mobilesal}, JLDCF \cite{fu2022siamese}, DENet \cite{xu2022weakly}, CCAFNet \cite{zhou2022ccafnet}, DWD \cite{zhang2022deep}, MoADNet \cite{jin2022moadnet}, MMNet \cite{gao2022unified}, EMANet \cite{feng2022encoder}, DCF \cite{sun2021deep}, CDINet \cite{zhang2021cross}, DFM-Net \cite{zhang2021depth}, BiANet \cite{zhang2021bilateral}, DQSD \cite{chen2021depth}, DRLF \cite{wang2021data}, cmSalGAN \cite{jiang2021cmsalgan}, and IRFRNet \cite{zhou2021irfr}. For fair comparisons, we compute their performance based on the saliency maps derived from their original papers. All the evaluation metrics are calculated by the official evaluation tools \cite{SIP_D3Net2021TNNLS}.

\begin{figure*}[!t]
\centering
\begin{minipage}[b]{0.49\linewidth}
    {\includegraphics[width=\linewidth]{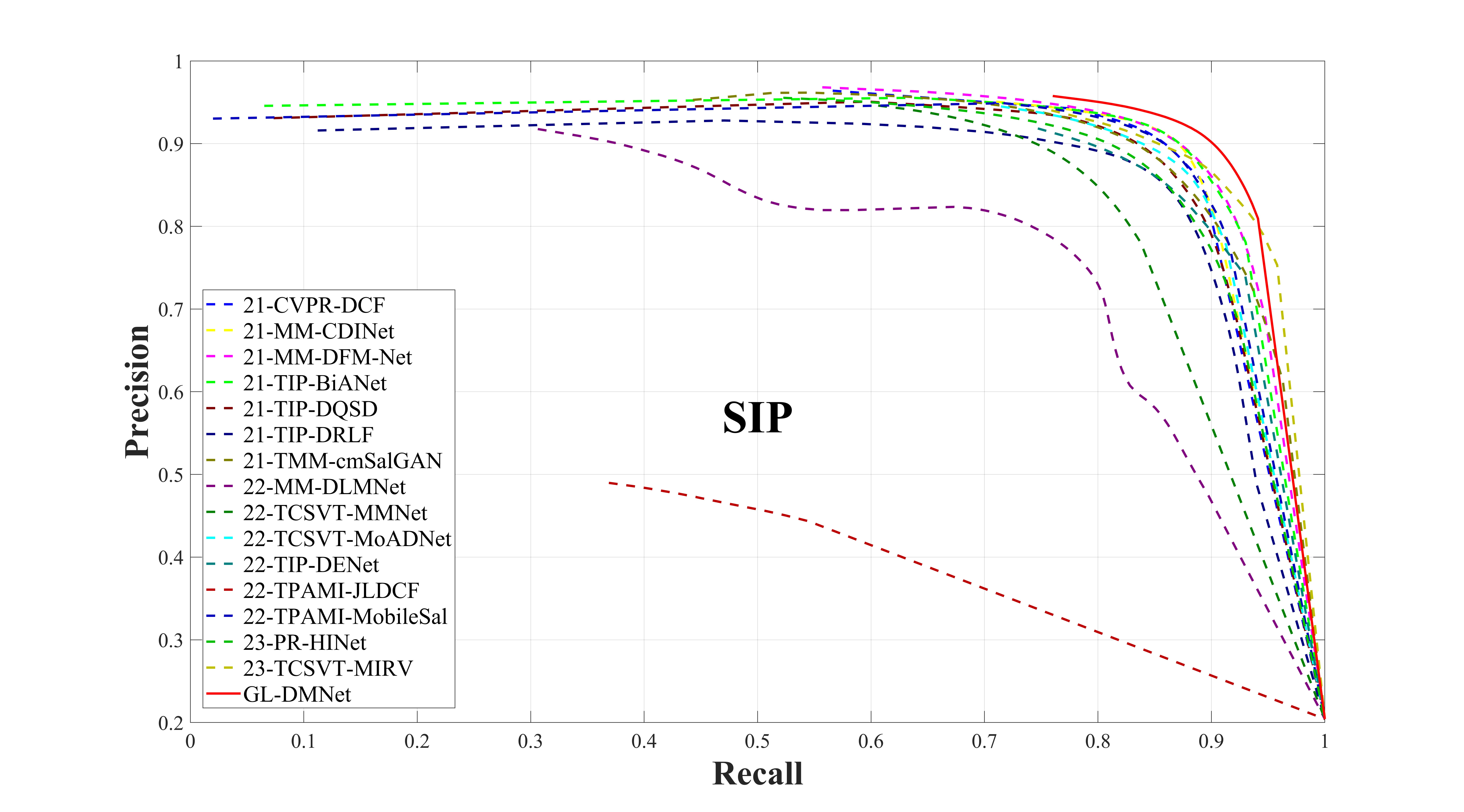}}
\end{minipage}
\begin{minipage}[b]{0.49\linewidth}
    {\includegraphics[width=\linewidth]{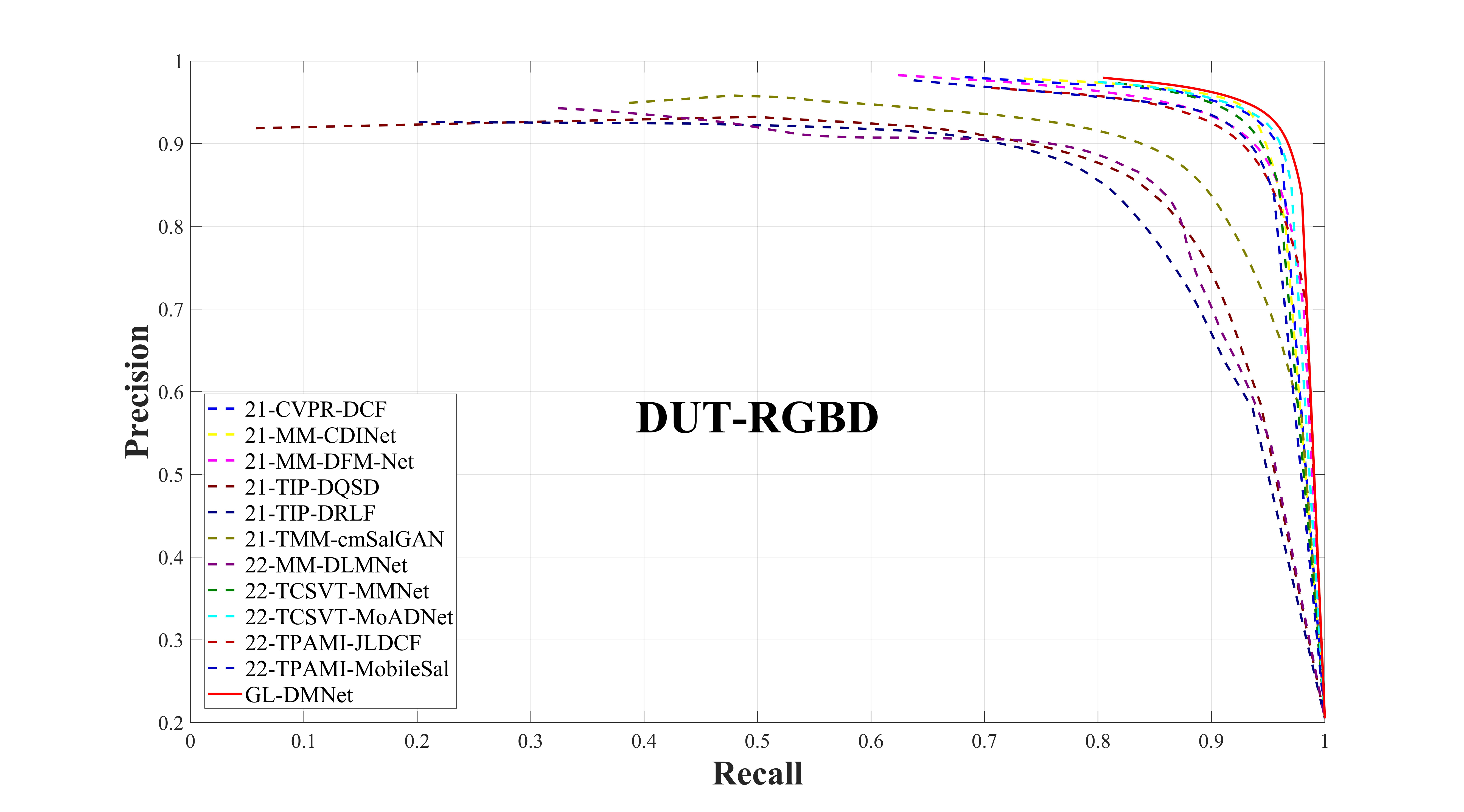}}
\end{minipage}
\hfill
\begin{minipage}[b]{0.49\linewidth}
    {\includegraphics[width=\linewidth]{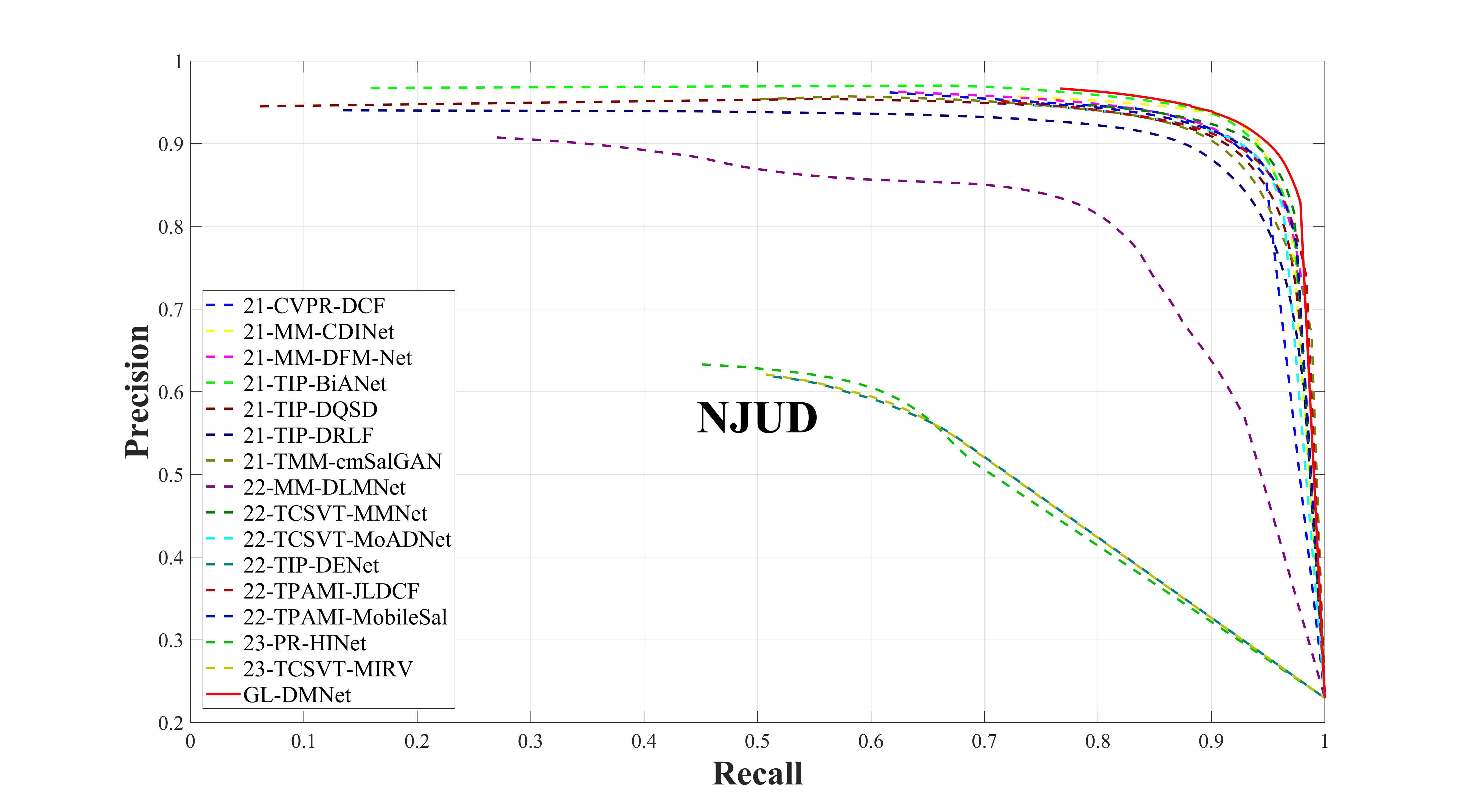}}
\end{minipage}
\begin{minipage}[b]{0.49\linewidth}
    {\includegraphics[width=\linewidth]{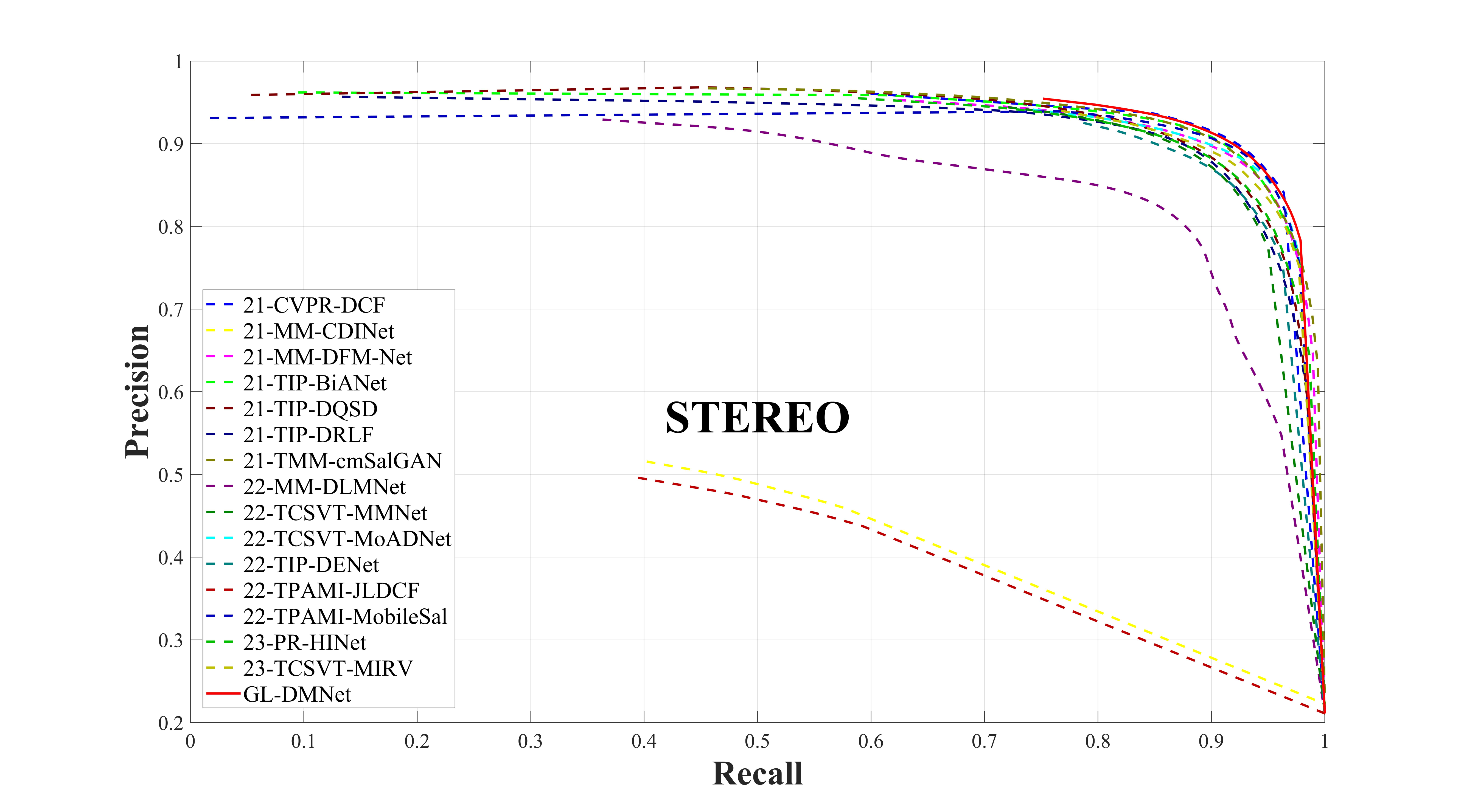}}
\end{minipage}
\hfill
\begin{minipage}[b]{0.49\linewidth}
    {\includegraphics[width=\linewidth]{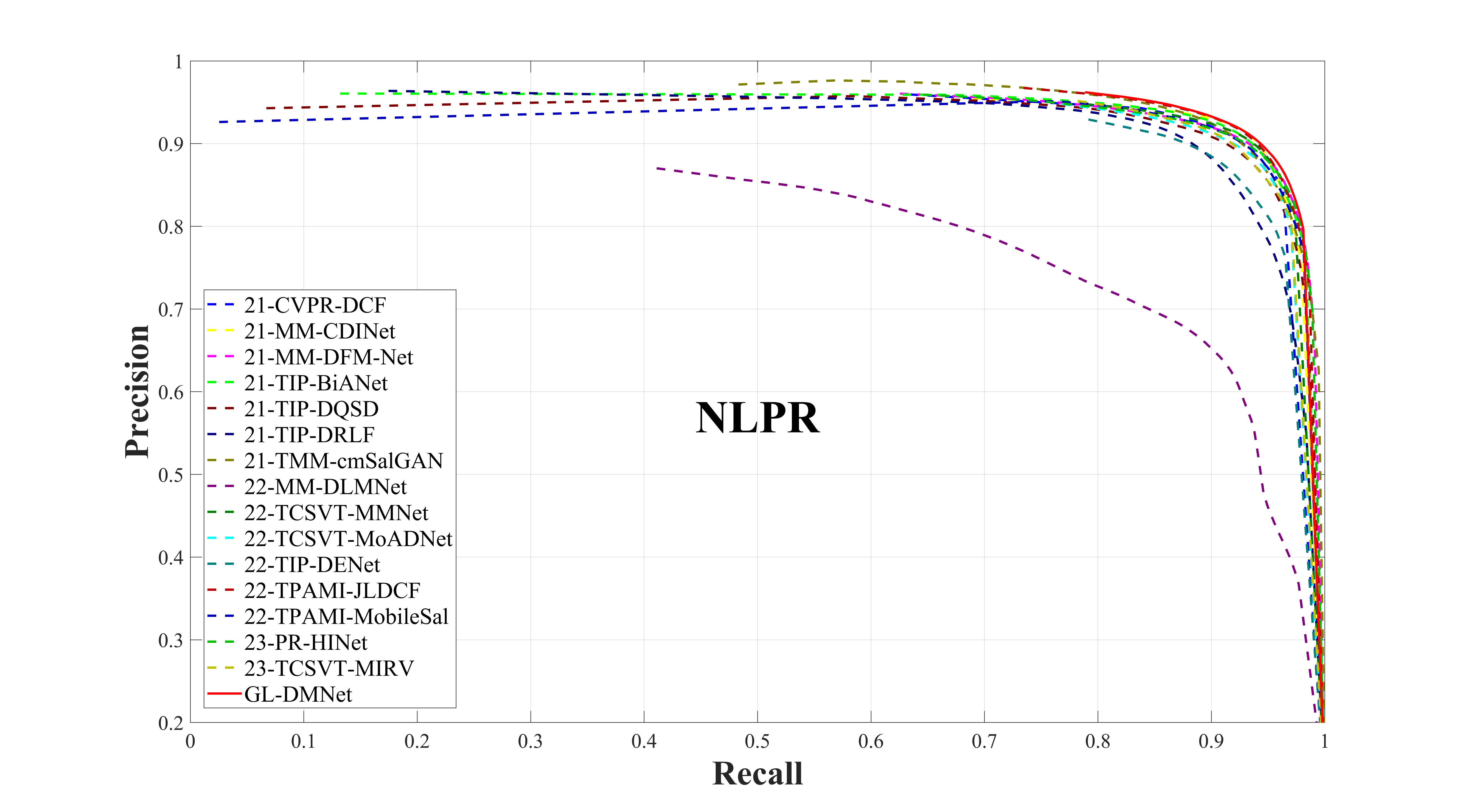}}
\end{minipage}
\begin{minipage}[b]{0.49\linewidth}
    {\includegraphics[width=\linewidth]{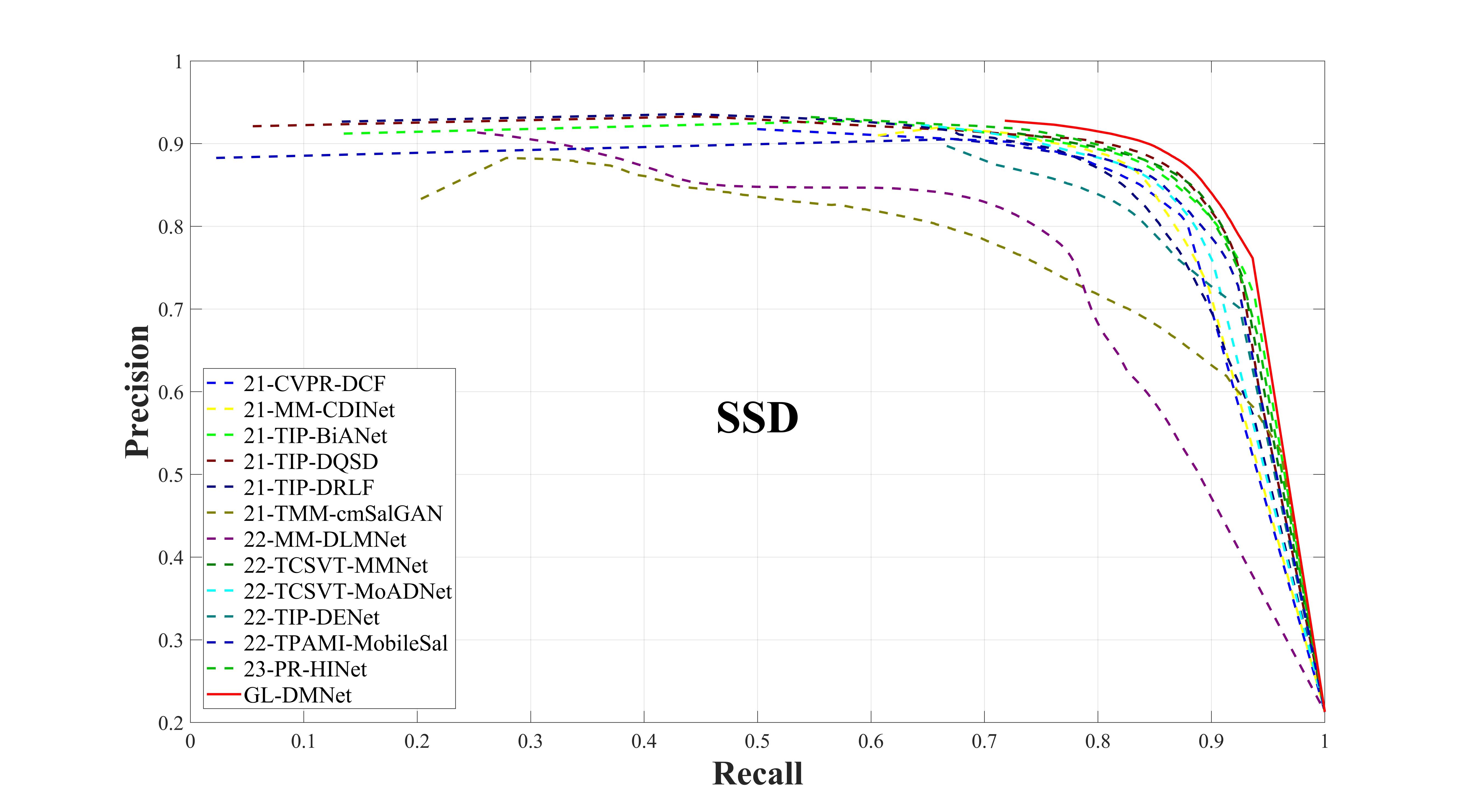}}
\end{minipage}
\caption{PR curves of different saliency detectors on six datasets.}
\label{fig:PR_curves}
\end{figure*}

\textbf{Quantitative Analysis:}
Table. \ref{tab:quantitative} illustrates the quantitative results of our model. We calculate four evaluation metrics on six datasets for comprehensive comparisons. Higher $S_{\alpha}$, $F_{\beta}$ and $E_{\xi}$ indicate better performance. On the contrary, lower ${\rm MAE}$ is better. We place other baselines in chronological order from top to bottom. Compared with other state-of-the-art methods, we could see that the proposed GL-DMNet almost achieves the best performance across all datasets, which demonstrates the effectiveness and superiority of our method. Besides, remarkable improvements are observed in all metrics, especially on the SIP and the DUT-RGBD datasets. The improvement gain of our model achieves 16.3\% of ${\rm MAE}$, 1.5\% of $S_{\alpha}$, 1.1\% of $F_{\beta}$ and 0.3\% of $E_{\xi}$ score on the SIP dataset when compared to the second best method. We also improve 3.3\% of ${\rm MAE}$, 0.5\% of $S_{\alpha}$, 0.3\% of $F_{\beta}$ and 0.6\% of $E_{\xi}$ score on the DUT-RGBD dataset. Moreover, FCFNet \cite{zhang2024feature} slightly outperforms our model in terms of the $E_{\xi}$ metric on the NJUD and STEREO datasets, which can be attributed to its strategy of discarding some low-quality depth maps. Additionally, S3Net \cite{zhu2023s} introduces supplementary prior information, which enables it to slightly surpass our model in the $S_{\alpha}$ and $F_{\beta}$ metrics on the STEREO dataset and the $F_{\beta}$ and ${\rm MAE}$ metrics on the NLPR dataset. Nevertheless, our proposed method outperforms theirs on the vast majority of datasets. We attribute our superior performance to the fusion strategies incorporating dual mutual learning, which places greater emphasis on detailed structures and contextual information across different modalities.

\textbf{Qualitative Analysis:} To further verify the performance of our proposed method, we provide some representative saliency maps predicted by GL-DMNet and other SOTA methods in Fig. \ref{fig:qualitative}. In particular, our method could capture more fine-grained details and pixel-level information in scenarios with intricate structures (1st row), exhibiting a holistic understanding of the localized features. Furthermore, in the cases of large objects (2nd row), small objects (3rd row), and multiple objects (4th-5th rows), our model efficiently and adequately distinguishes the outline of the object from the background, which indicates its strong robustness in discerning salient objects with varying scales. Additionally, even though depth information is absent (6th-7th rows), our model could still extract meaningful cues from 2D visual information alone and produce accurate saliency maps. Our model also performs better when dealing with complex scenes (8th-9th rows) and low contrast (10th row). These results validate the superiority of our model in handling different visual environments, with consistent and satisfactory performance for multiple challenging scenarios.

\begin{table*}[!t]
\centering
\caption{Ablation experiment of dual mutual fusion module. The best result is in bold.}
\label{tab:ablation-DMF}
\begin{tabular}{c|cccc|cccc|cccc}
\hline
\multirow{2}{*}{Variant} & \multicolumn{4}{c|}{Candidate} & \multicolumn{4}{c|}{NLPR} & \multicolumn{4}{c}{DUT-RGBD} \\
 & PMF & CMF & Serial & Parallel & $E_\xi$ $\uparrow$ & $S_\alpha$ $\uparrow$ & $F_\beta$ $\uparrow$ & MAE $\downarrow$ & $E_\xi$ $\uparrow$ & $S_\alpha$ $\uparrow$ & $F_\beta$ $\uparrow$ & MAE $\downarrow$ \\ \hline
No.1 & \checkmark &  &  &  & 0.955 & 0.920 & 0.915 & 0.026 & 0.958 & 0.924 & 0.940 & 0.033 \\
No.2 &  & \checkmark &  &  & 0.957 & 0.922 & 0.921 & 0.026 & 0.960 & 0.930 & 0.946 & 0.030 \\
No.3 & \checkmark & \checkmark & \checkmark &  & 0.960 & \textbf{0.927} & 0.925 & 0.023 & 0.959 & 0.925 & 0.942 & 0.033 \\
No.4 &  &  &  & \checkmark & 0.952 & 0.920 & 0.915 & 0.026 & 0.955 & 0.921 & 0.939 & 0.033 \\
No.5 & \checkmark & \checkmark &  & \checkmark & \textbf{0.962} & \textbf{0.927} & \textbf{0.926} & \textbf{0.022} & \textbf{0.962} & \textbf{0.931} & \textbf{0.947} & \textbf{0.029} \\ \hline
\end{tabular}
\end{table*}

\begin{table*}[!t]
\centering
\caption{Ablation experiment of cascade transformer-infused reconstruction decoder. The best result is in bold.}
\label{tab:ablation-Decoder}
\begin{tabular}{c|ccc|cccc|cccc}
\hline
\multirow{2}{*}{Variant} & \multicolumn{3}{c|}{Candidate} & \multicolumn{4}{c|}{NLPR} & \multicolumn{4}{c}{DUT-RGBD} \\
 & PVT V2 & PVT V1 & Reconstruction & $E_\xi$ $\uparrow$ & $S_\alpha$ $\uparrow$ & $F_\beta$ $\uparrow$ & MAE $\downarrow$ & $E_\xi$ $\uparrow$ & $S_\alpha$ $\uparrow$ & $F_\beta$ $\uparrow$ & MAE $\downarrow$ \\ \hline
No.1 & \checkmark &  &  & 0.954 & 0.921 & 0.921 & 0.024 & 0.959 & 0.928 & 0.945 & 0.030 \\
No.2 &  & \checkmark &  & 0.948 & 0.918 & 0.919 & 0.027 & 0.958 & 0.928 & 0.943 & 0.032 \\
No.3 &  &  & \checkmark & 0.951 & 0.917 & 0.916 & 0.028 & 0.955 & 0.925 & 0.939 & 0.034 \\
No.4 &  & \checkmark & \checkmark & 0.958 & 0.921 & 0.918 & 0.025 & 0.961 & \textbf{0.931} & 0.945 & 0.030 \\
No.5 & \checkmark &  & \checkmark & \textbf{0.962} & \textbf{0.927} & \textbf{0.926} & \textbf{0.022} & \textbf{0.962} & \textbf{0.931} & \textbf{0.947} & \textbf{0.029} \\ \hline
\end{tabular}
\end{table*}

\textbf{PR Curves:} As shown in Fig. \ref{fig:PR_curves}, the PR curves visually depict the comparison between the proposed GL-DMNet and other state-of-the-art methods on six RGB-D datasets. The red solid line indicates that our method outperforms all compared methods across most threshold values.

\subsection{Ablation studies}
\begin{figure}[ht]
	\centering
	\includegraphics[width=0.48\textwidth]{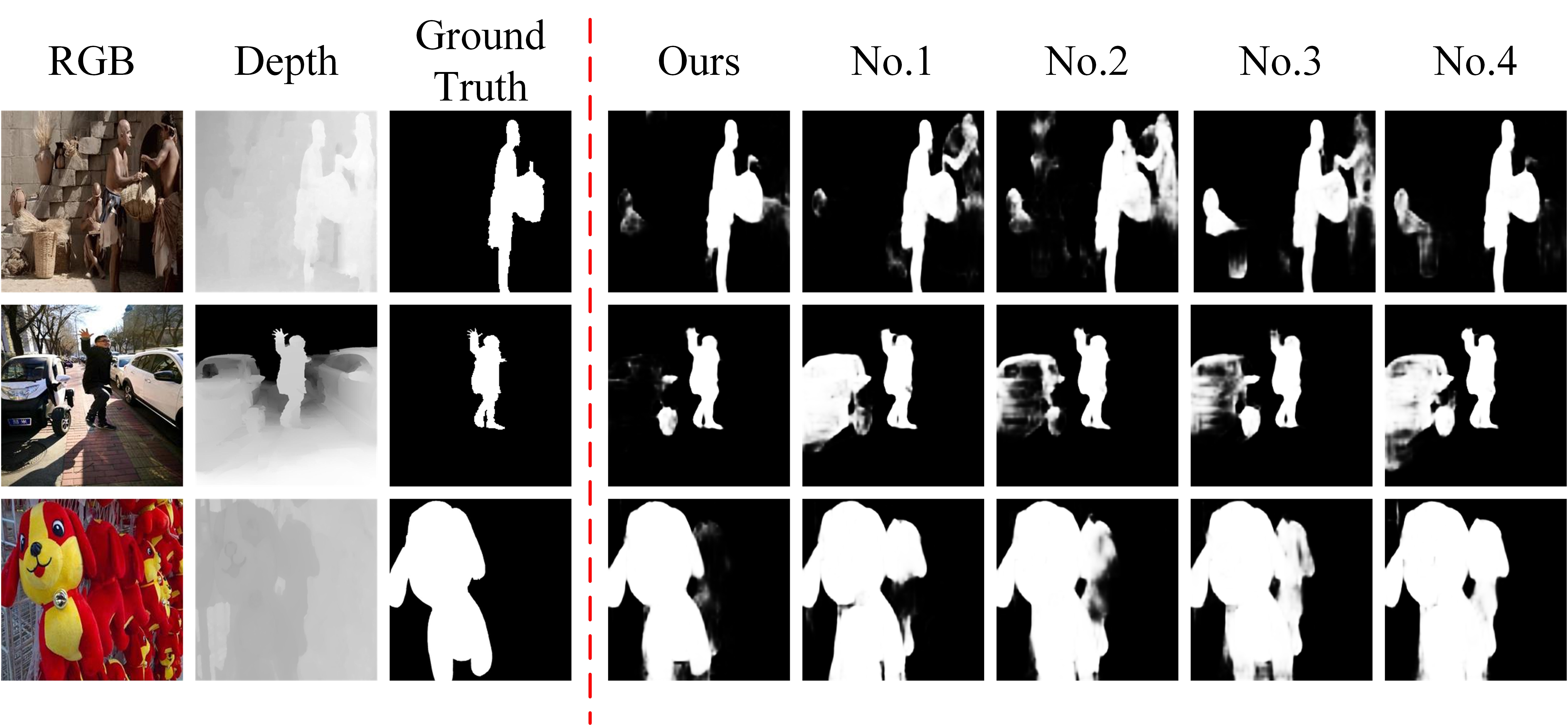}
	\centering
	\caption{Visual comparison of saliency map results produced by different variants of fusion modules. Please refer to Table \ref{tab:ablation-DMF} for the explanation of No.1 to No.4.}
	\label{fig:ablation-fusion}
\end{figure}

\begin{figure}[ht]
	\centering
	\includegraphics[width=0.48\textwidth]{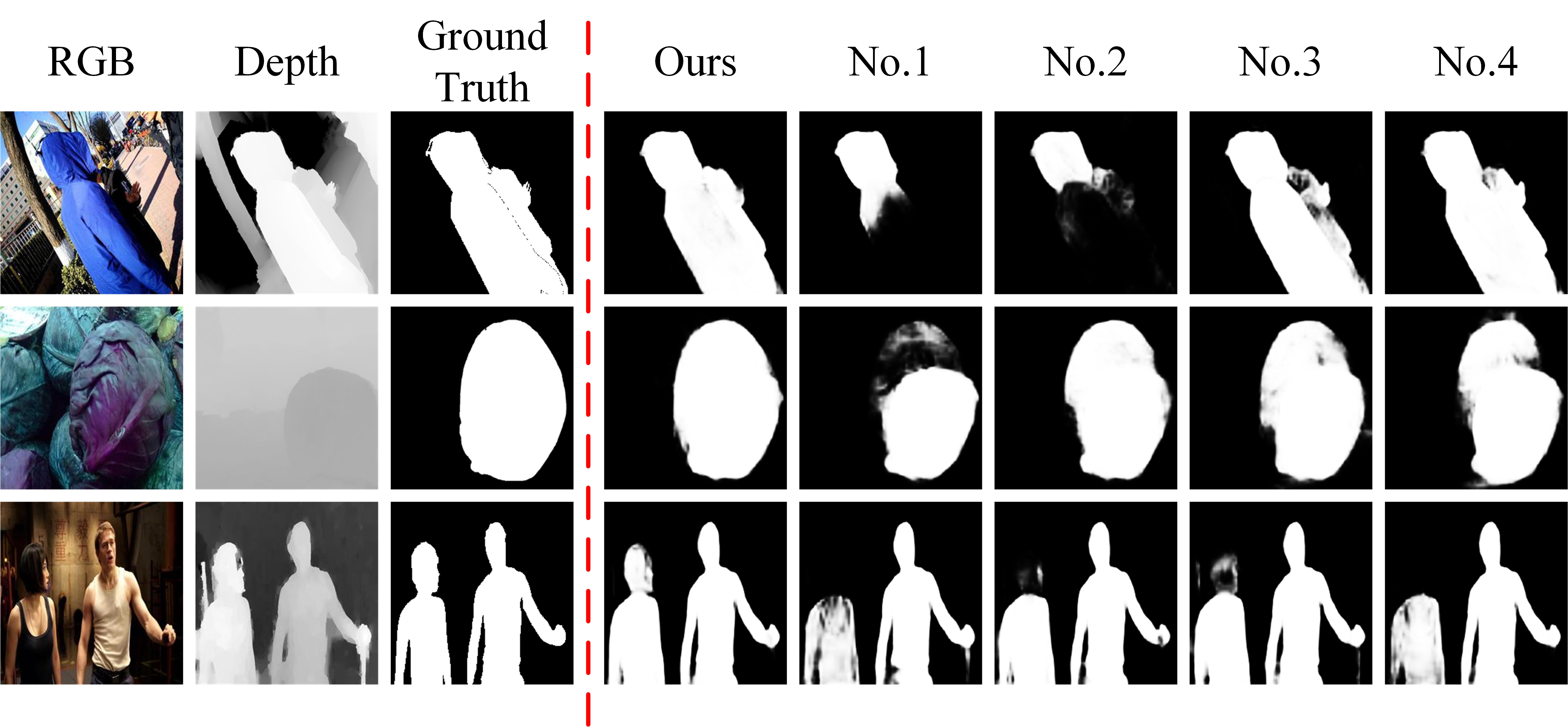}
	\centering
	\caption{Visual comparison of saliency map results produced by different variants of decoder. Please refer to Table \ref{tab:ablation-Decoder} for the explanation of No.1 to No.4.}
	\label{fig:ablation-decoder}
\end{figure}

To further investigate the effectiveness of each key component in the proposed GL-DMNet, we conduct ablation studies on NLPR and DUT-RGBD datasets by systematically removing components or replacing them with similar structures to assess their impact on GL-DMNet. Fig. \ref{fig:ablation-fusion} and Fig. \ref{fig:ablation-decoder} also show the qualitative comparisons of different variants.

\textbf{The effectiveness of dual mutual fusion module.}
Table \ref{tab:ablation-DMF} illustrates the ablation results on the dual-mutual fusion module. We consider four variants of fusion strategies: (1) only PMF, (2) only CMF, (3) serial utilization of PMF and CMF, (4) direct connection without other operations, and (5) our dual mutual design. First of all, it is observed that directly catenating multi-modal features without learning their independence would harm the performance, which is the worst among all the ablative variants. Second,  adopting PMF (from spatial dimension) or CMF (from channel dimension) can only bring limited improvement. This may suggest that it is necessary to combine of these two types for better performance. Thus, if we use the serial means for them, the performance could be further improved. However, our dual mutual that utilizes both spatial and channel dimensions to learn the deep correlation and independence, outperforms all the variants. The dual mutual design promotes the interactions between cross-modality features. Hence, it becomes possible to fuse multi-modal embeddings without destroying their individual characteristics.

\textbf{The effectiveness of transformer embedding.}
In our GL-DMNet, we deploy Transformer-induced architecture to learn the global representations of the multi-level fusion features. Now we verify the different choice of the Transformer-based embedding: (1) PVTv2 and (2) PVTv1. The results are shown in the first and second rows of Table \ref{tab:ablation-Decoder}. Obviously, the performance of PVTv2 is better than PVTv1 over all the evaluation metrics. Compared with the PVTv1, PVTv2 could capture global information while protecting local continuity at the same time. Thus, PVTv2 is apparently more suitable for our model where we aim to explore global-local awareness for RGB-D SOD task. Moreover, PVTv2 is capable of speeding up the decoding process of Transformer-induced architecture.

\textbf{The effectiveness of multi-level feature reconstruction.} To validate the effectiveness of the adopted reconstruction decoder for the multi-level features, we conduct comparative experiments on the ablated types: (1) only feature reconstruction without Transformer-based embedding (e.g., PVTv2 and PVTv1), (2) reconstruction on the output from PVTv2, and (3) reconstruction on the output from PVTv1. The results are presented in the fourth and fiveth rows of Table \ref{tab:ablation-Decoder}. removing the Transformer-based feature learning is not beneficial to the reconstruction decoding, leading to a worse performance after feature reconstruction. This is because there exists inherent discrepancy between difference levels. Transformer-based embedding will help the reconstruction to understand it and decode them more efficiently. Thus, the performance of reconstruction with PVTv2 or PVTv1 gets improved. However, compared with PVTv1-based reconstruction, the reconstruction with PVTv2 is much superior, obtaining more improvement. This indicates the global-local awareness captured by PVTv2 would further strengthen the decoding process of multi-level feature reconstruction. Hence, we eventually decide to adopt the reconstruction decoder with PVTv2.

\begin{table}[ht]
    \centering
    \caption{Ablation experiment of loss function. The best result is in bold.}
    \label{tab:ablation-loss}
\begin{tabular}{c@{\hskip 2pt}|c@{\hskip 2pt}c@{\hskip 2pt}c@{\hskip 2pt}c@{\hskip 2pt}|c@{\hskip 2pt}c@{\hskip 2pt}c@{\hskip 2pt}c@{\hskip 1pt}}
\hline
\multirow{2}{*}{Loss} & \multicolumn{4}{c|}{NJUD} & \multicolumn{4}{c}{SIP} \\ \cline{2-9} 
 & $E_\xi$ $\uparrow$ & $S_\alpha$ $\uparrow$ & $F_\beta$ $\uparrow$ & MAE $\downarrow$ & $E_\xi$ $\uparrow$ & $S_\alpha$ $\uparrow$ & $F_\beta$ $\uparrow$ & MAE $\downarrow$ \\ \hline
BCE & 0.941 & 0.918 & 0.928 & 0.038 & 0.927 & 0.888 & 0.907 & 0.048 \\
BCE+DICE & 0.939 & 0.919 & 0.926 & 0.038 & 0.930 & 0.895 & 0.912 & 0.045 \\
BCE+SSIM & 0.943 & 0.917 & 0.926 & 0.035 & 0.927 & 0.888 & 0.910 & 0.045 \\
Ours & \textbf{0.950} & \textbf{0.920} & \textbf{0.930} & \textbf{0.033} & \textbf{0.936} & \textbf{0.896} & \textbf{0.915} & \textbf{0.041} \\ \hline
\end{tabular}
\end{table}

\textbf{The effectiveness of loss function.} To demonstrate the effectiveness and optimality of our loss function design, we conducted experiments using three different combinations of loss functions for comparison. The results show that our combination consistently outperforms the other three on the NJUD and SIP datasets, thereby establishing the superiority of our design. The results in the table \ref{tab:ablation-loss} highlight the robust performance of our loss function.

\subsection{Visualization}
\begin{figure}[ht]
\centering
\includegraphics[width=0.45\textwidth]{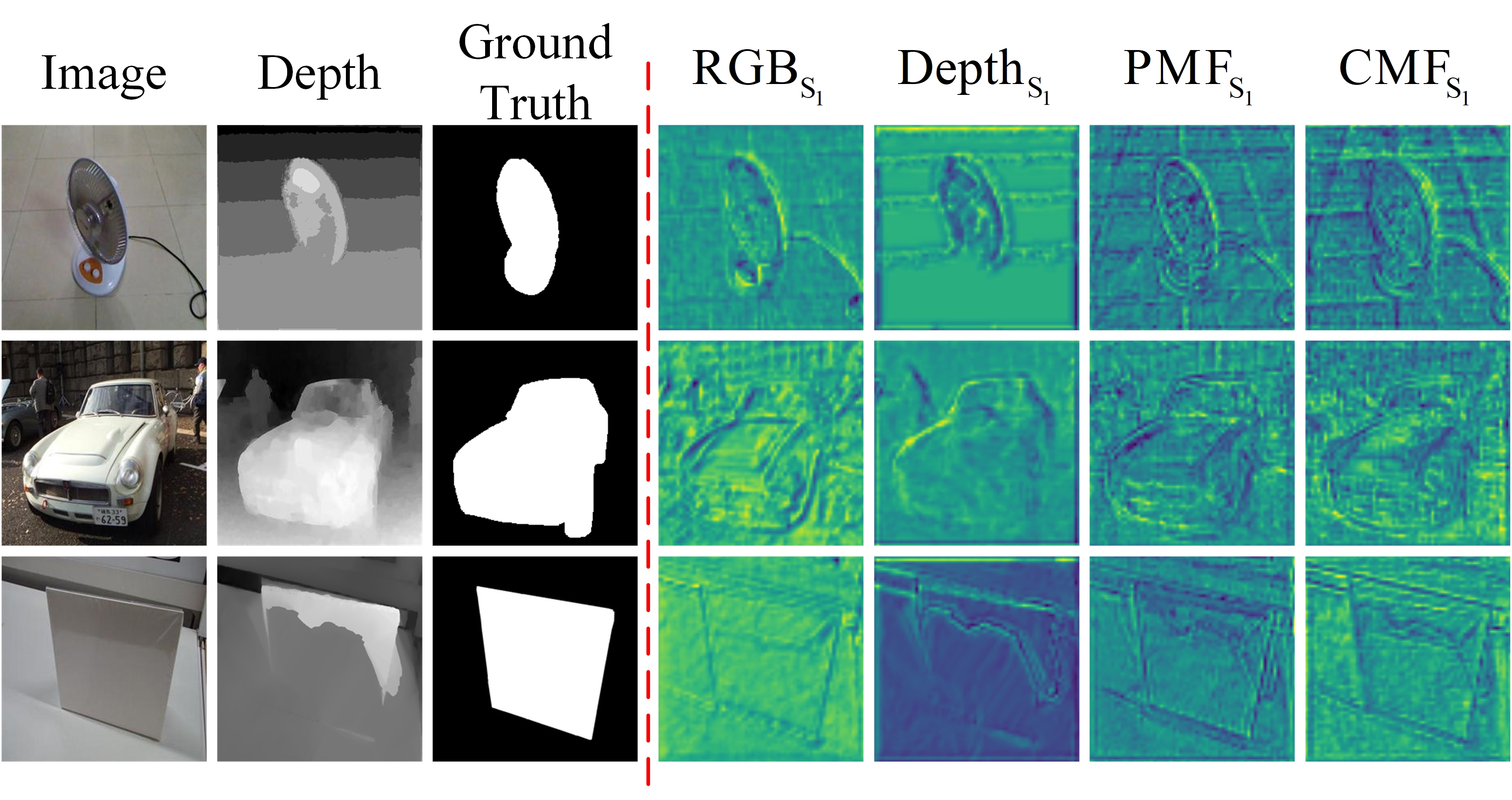}
\caption{Feature visualizations of the PMF and CMF. $\mathrm S_1$ represents the channel images selected from the fusion module of Stage 1.}
\label{fig:Visualization}
\end{figure}

Fig. \ref{fig:Visualization} illustrates the separated channel images processed by the PMF and CMF modules. Compared to the RGB and depth images of the same channels, the images processed by the PMF and CMF modules exhibit clearer edge contours and richer details. This enhancement addresses the inherent discrepancies between RGB and depth information. Furthermore, it demonstrates the effectiveness of the PMF and CMF modules, which fully leverage the interrelationships between different modalities in spatial and channel dimensions, thereby enhancing the feature fusion effect.

\section{Conclusion and future work}
In this work, we propose a novel model called dual mutual learning network with global–local awareness (GL-DMNet) for the RGB-D salient object detection task. Our motivation is existing methods directly fuse attentional cross-modality features under manual-mandatory fusion strategy without considering the inherent discrepancy between the RGB and depth. Firstly, we present the position mutual fusion and channel mutual fusion modules by parallel design to efficiently extract features. Secondly, to distinguish the discrepancy and independence across multi-level features, we carefully design the cascade transformer-infused reconstruction for multi-level feature decoding. Last, we conduct extensive experiments on six public datasets over 24 state-of-the-art baselines. The experimental results demonstrate the superiority and effectiveness of our GL-DMNet. In the future, we intend to leverage the proposed model to tackle several practical challenges within the medical imaging domain, aiming to enhance the model’s generalization ability and robustness and broaden its applicability. By developing a universal and effective multimodal model, we aspire to simultaneously process data from diverse modalities such as RGB, depth, and thermal. This approach is expected to significantly enhance the model’s practical utility, enabling it to integrate and analyze complex datasets and thus provide more comprehensive insights into various medical conditions.


\bibliographystyle{IEEEtran}
\bibliography{reference}

\begin{IEEEbiography}[{\includegraphics[width=1in,height=1.25in,clip,keepaspectratio]{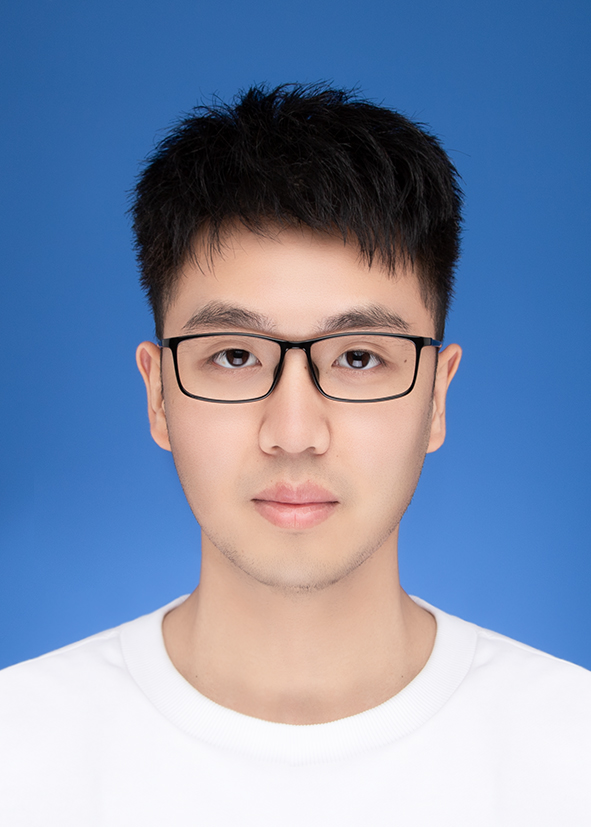}}]{Kang Yi}
    received the B.S. degree from China Agricultural University, Beijing, China, in computer science. He is currently pursuing his Ph.D. degree with College of Artificial Intelligence from Nankai University, Tianjin, China. In addition, he had been a visiting scholar in the Polytechnic University of Hong Kong. His research interests include computer vision and multimedia computing.
\end{IEEEbiography}

\begin{IEEEbiography}[{\includegraphics[width=1in,height=1.25in,clip,keepaspectratio]{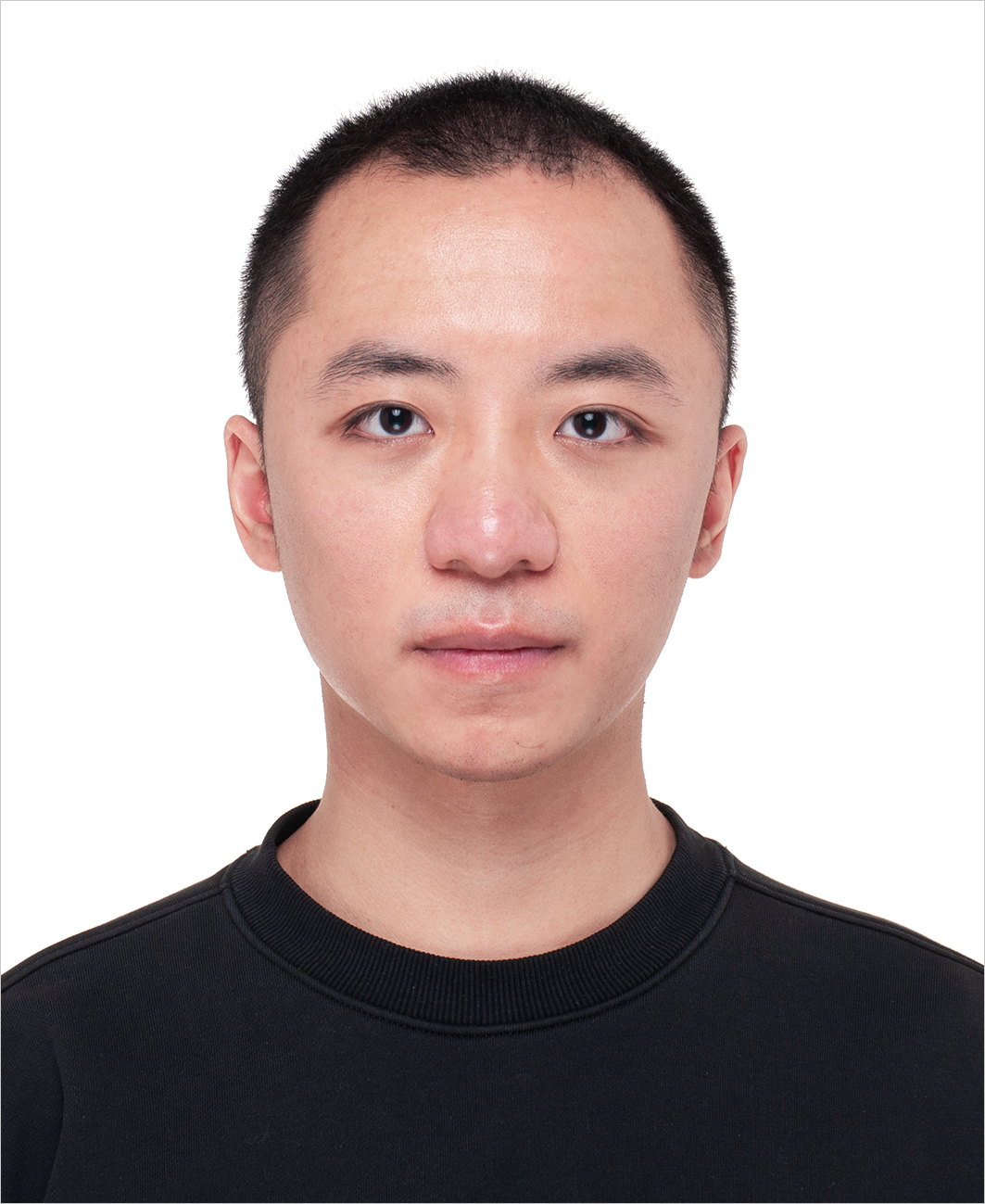}}]{Haoran Tang} 
    is now a Ph.D. student at the Hong Kong Polytechnic University and the Univerity of Technology Sydney. He received B.S. and M.S. degrees from the Chongqing University. His research interests include temporal graph learning and data mining.
\end{IEEEbiography}

\begin{IEEEbiography}[{\includegraphics[width=1in,height=1.25in,clip,keepaspectratio]{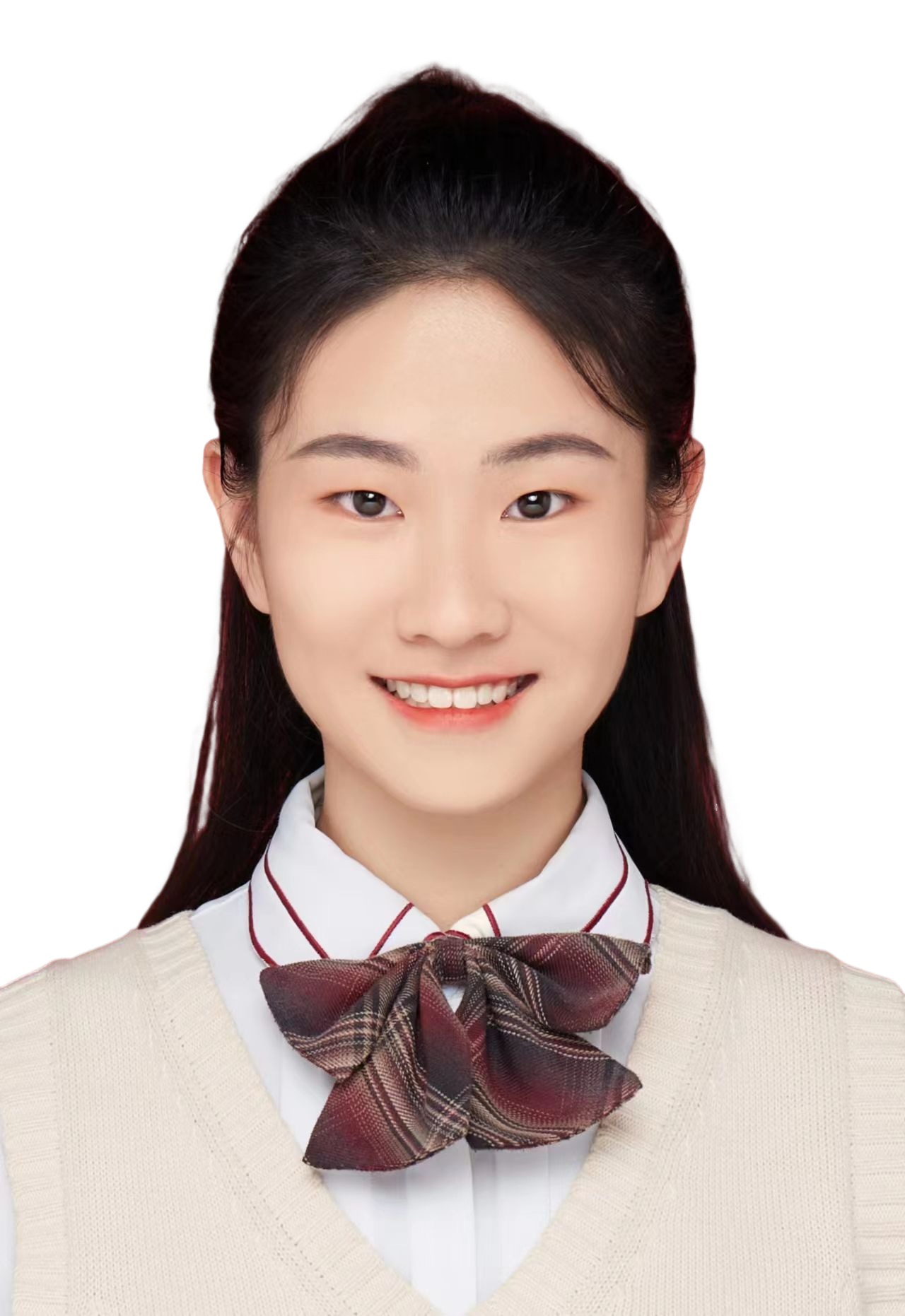}}]{Yumeng Li}
    is studying at the College of Artificial Intelligence, Nankai University. Her research interests include computer vision and medical image processing. 
\end{IEEEbiography}

\begin{IEEEbiography}[{\includegraphics[width=1in,height=1.25in,clip,keepaspectratio]{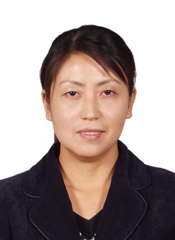}}]{Jing Xu}
    is a professor at the College of Artificial Intelligence, Nankai University. She received her Ph.D. degree from Nankai University in 2003. She has published more than 100 papers in software engineering, software security, and big data analytics. She won the second prize of the Tianjin Science and Technology Progress Award twice in 2017 and 2018, respectively.
\end{IEEEbiography}

\begin{IEEEbiography}[{\includegraphics[width=1in,height=1.25in,clip,keepaspectratio]{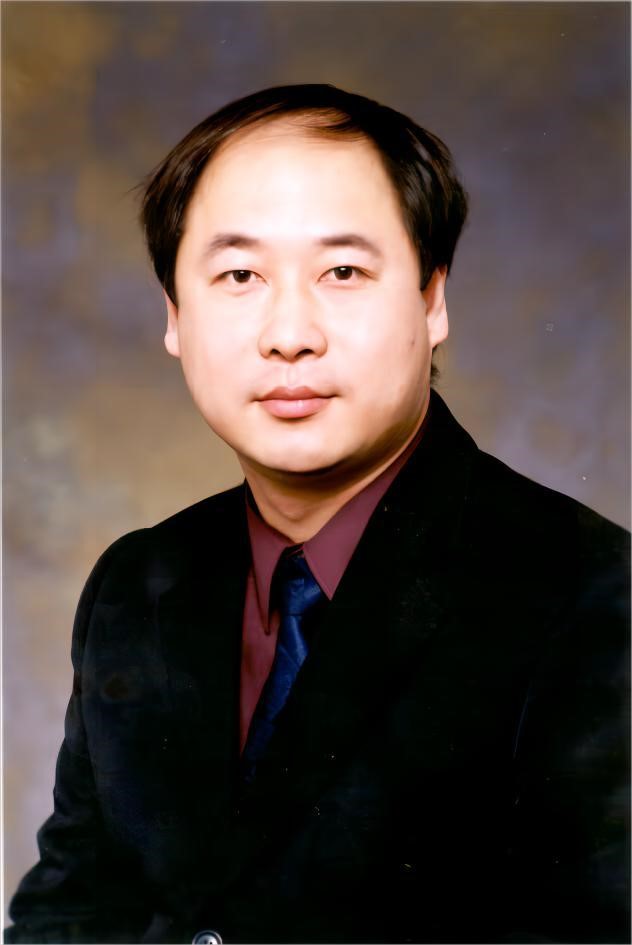}}]{Jun Zhang}
    received his PhD degree in Electrical Engineering from the City University of Hong Kong in 2002. His research activities are mainly in the areas of computational intelligence. Based on his research in evolutionary computation and its applications, Zhang Jun has published more than 600 peer-reviewed research papers, of which more than 200 have been published in IEEE Transactions. He currently serves as Associate Editor of IEEE Transactions on Artificial Intelligence and IEEE Transactions on Cybernetics.
\end{IEEEbiography}

\vfill
\end{document}